%% file: neurips_2026.tex
\documentclass{article}

\usepackage[main, final]{neurips_2026}

\usepackage[utf8]{inputenc} 
\usepackage[T1]{fontenc}    
\usepackage{hyperref}       
\usepackage{url}            
\usepackage{booktabs}       
\usepackage{amsfonts}       
\usepackage{nicefrac}       
\usepackage{microtype}      
\usepackage[table]{xcolor}         

\usepackage{graphicx}
\usepackage{multibib}
\usepackage{caption}
\usepackage{multirow}
\ifdefined\pdfcompresslevel
\usepackage[accsupp]{axessibility}
\fi
\usepackage{amsthm,amsmath,amssymb}

\theoremstyle{plain}

\theoremstyle{definition}

\theoremstyle{remark}

\newcommand{\eg}{{\emph{e.g.}}, }
\newcommand{\etc}{etc.}
\newcommand{\ie}{{\emph{i.e.}}, }
\definecolor{myblue}{RGB}{212, 239, 251}
\definecolor{mygreen}{RGB}{217, 255, 217}
\definecolor{myyellow}{RGB}{255, 252, 200}

\definecolor{baselinecolor}{gray}{.9}

\input{preamble}

\title{LLM Is a Good Conditioner: End-to-End Sign Language Video Generation with VQ-Diffusion} 

\author{%
  Xiao Liu \\
  State Key Laboratory of Novel Software \\Technology \\
  Nanjing University \\
  Suzhou, China \\
  \texttt{liuxiaox@smail.nju.edu.cn} \\
  \And
  Shiwei Gan\thanks{Corresponding author.} \\
  State Key Laboratory of Novel Software \\Technology \\
  Nanjing University \\
  Nanjing, China \\
  \texttt{sw@nju.edu.cn} \\
}

\begin{document}

\maketitle

{\vspace{-3mm}
\centering
 \captionsetup{type=figure}
 \includegraphics[width=1\textwidth]{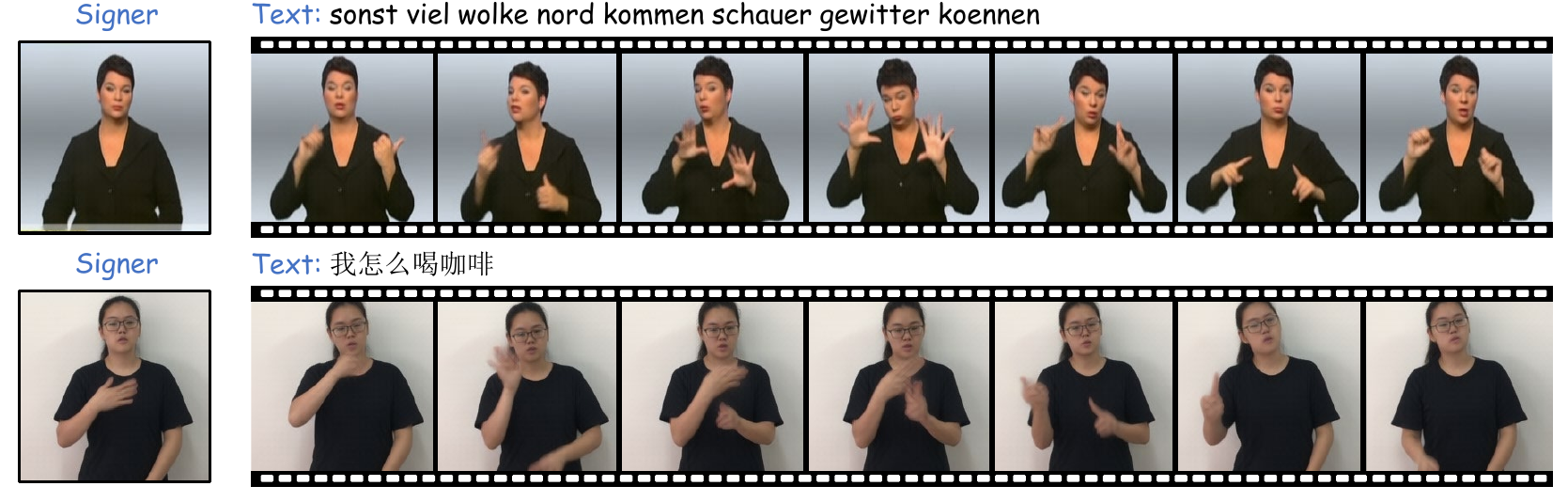}
 \vspace{-7mm}
 \captionof{figure}{Generated sign language videos from our SLVG framework. Given a spoken-language sentence and a single reference frame, the model generates semantically aligned and temporally smooth sign videos without the need for intermediate cues (e.g., pose or hands).} 
 \label{fig:teaser}
 \par}

 \input{sec/0_abstract}

 \input{sec/1_intro}

 \input{sec/2_related}

 \input{sec/3_method}

 \input{sec/4_experiment}

\input{sec/4.1_compare}

 \input{sec/5_performance}

\input{sec/6_conclusion} 

\section{Acknowledgments}
This work was supported by the National Natural Science Foundation of China (NSFC) under Grant No. 62606221, the Postdoctoral Fellowship Program of the China Postdoctoral Science Foundation (CPSF) under Grant No. GZB20260384, and the Jiangsu Funding Program for Excellent Postdoctoral Talent under Grant No. 2026ZB254.

\bibliographystyle{plainnat}
\bibliography{bib/new, bib/cslr, bib/dataset, bib/islr, bib/other, bib/slt}


\newpage
\appendix


\input{sec/sup}

\clearpage
\newpage
\input{checklist.tex}

\end{document}

%% file: preamble.tex
\usepackage{graphicx} 
\usepackage{booktabs}
\usepackage{multibib}
\usepackage{pifont}

\usepackage{caption}
\usepackage{subcaption}
\usepackage{multirow}  
\usepackage{wrapfig}
\usepackage{amsthm,amsmath,amssymb}

\usepackage{mathrsfs}
\usepackage{bbding}
\usepackage{url}            %
\usepackage[table]{xcolor} 

\definecolor{LightBlue}{RGB}{240,248,255}
\definecolor{deepgray}{gray}{0.3}

\newlength\savewidth
\newcommand{\tablestyle}[2]{\setlength{\tabcolsep}{#1}\renewcommand{\arraystretch}{#2}\centering\footnotesize}

\makeatletter\renewcommand\paragraph{\@startsection{paragraph}{4}{\z@}
  {.5em \@plus1ex \@minus.2ex}{-.5em}{\normalfont\normalsize\bfseries}}\makeatother

\newcolumntype{x}[1]{>{\centering\arraybackslash}p{#1pt}}
\newcolumntype{y}[1]{>{\raggedright\arraybackslash}p{#1pt}}
\newcolumntype{z}[1]{>{\raggedleft\arraybackslash}p{#1pt}}

\newcommand{\app}{\raise.17ex\hbox{$\scriptstyle\sim$}}

\definecolor{baselinecolor}{gray}{.9}
\newcommand{\baseline}[1]{\cellcolor{baselinecolor}{#1}}

\newcolumntype{*}{>{\global\let\currentrowstyle\relax}}
\newcolumntype{^}{>{\currentrowstyle}}

\definecolor{dt}{gray}{0.7}  %

\usepackage[capitalize]{cleveref}
\crefname{section}{Sec.}{Secs.}
\Crefname{section}{Section}{Sections}
\Crefname{table}{Table}{Tables}
\crefname{table}{Tab.}{Tabs.}

\newcolumntype{S}{@{}>{\lrbox0}l<{\endlrbox}}  %
\definecolor{lightgreen}{HTML}{D8ECD1}

\usepackage{tikz}

\let\cite\citep

%% file: sec/0_abstract.tex
\begin{abstract}
\label{sec:abstract}
Sign Language Video Generation (SLVG) aims to generate realistic and motion-accurate sign language videos from spoken text. Most existing studies heavily rely on intermediate representations (\eg poses or 3D meshes), and adopt a two-stage generation paradigm, where the text is first converted into poses or other visual modalities, which are then used as conditions to synthesize sign language video frames via diffusion models (\eg pose-to-image models). 
However, the end-to-end SLVG task remains largely unexplored.  In this paper, we demonstrate that end-to-end SLVG is feasible without relying on intermediate representations (\eg poses), and we argue that a key challenge lies in how to generate effective and temporally consistent conditions. With well-designed condition modules, it becomes feasible to build an end-to-end SLVG model without relying on any intermediate modalities. To this end, we propose LLMCond, a first end-to-end text-to-video SLVG framework that requires no auxiliary visual cues during either training or inference.
LLMCond consists of four key components:
(1) a video VQ-VAE that compresses sign language videos into a sequence of discrete latent tokens;
(2) \textbf{an LLM-based conditioner} that generates temporally and semantically aligned conditions corresponding to the sign sequences;
(3) \textbf{a Gaussian Condition Refiner} designed to enforce temporal smoothness across conditions, enabling the generation of coherent and natural sign language motions; and
(4) a discrete diffusion model that synthesizes motion-accurate sign language videos conditioned on the refined condition sequences. Extensive experiments on public SL datasets demonstrate that LLMCond achieves highly competitive performance, producing temporally coherent and accurate sign language motions without relying on auxiliary visual modalities such as pose, depth, or optical flow.
\end{abstract}
\vspace{-5mm}

%% file: sec/1_intro.tex
\section{Introduction}
\label{sec:intrduction}
\label{sec:intro}

\begin{figure*}[t]
  \centering
    \includegraphics[width=1.0\linewidth]{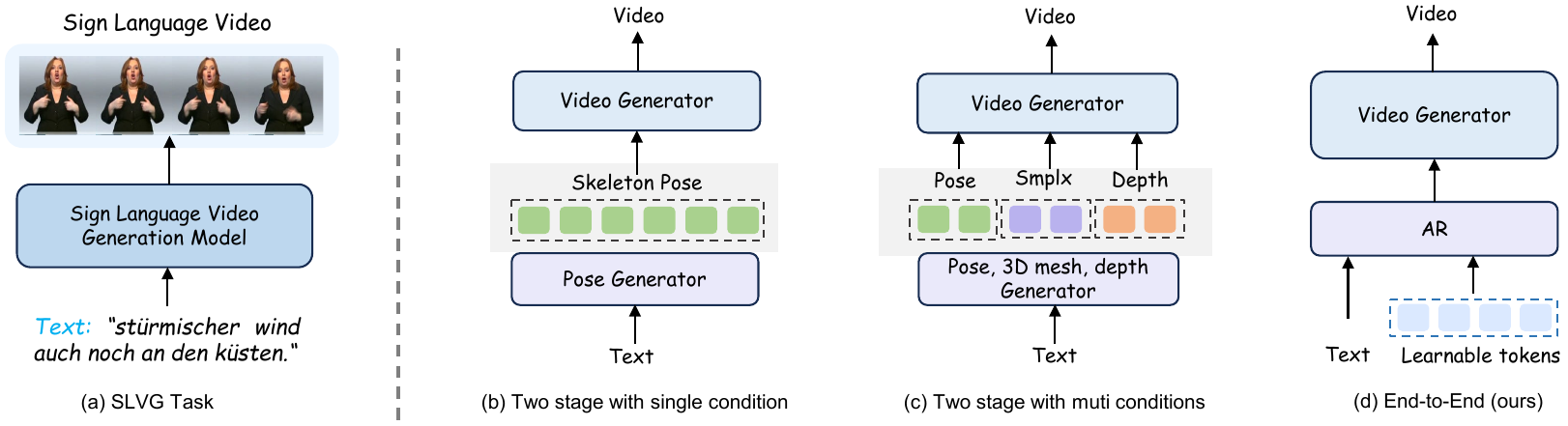}
    \vspace{-1mm}
   \caption{Difference between two-stage SLVG models and our end-to-end approach. Our model directly learns conditional tokens from text and generates videos without relying on intermediate modalities (pose, 3D mesh, depth).}
   \label{fig:motivation} 
  \vspace{-4mm}
\end{figure*}

Research on sign language (SL) aims to enable bidirectional conversion between SL and spoken language. Accordingly, it is typically divided into two main tasks: Sign Language Understanding and Sign Language Generation. The former focuses on translating sign language sequences into glosses or textual sentences, which encompasses various sub-tasks such as Isolated SL Recognition (ISLR)~\cite{hu2021hand,huang2018attention}, Continuous SL Recognition (CSLR)~\cite{gan2024signgraph,wei2023improving},  Sign Language Translation (SLT)~\cite{gan2026signllama,chen2022two} and SL Question Answering (SLQA)~\cite{gan2026sign}. The latter, in contrast, aims to generate corresponding sign pose or video sequences from text sequences, leading to tasks such as Sign Language Pose Production (SLP)~\cite{saunders2021continuous,baltatzis2024neural,liu2026signpr}, Sign Language Video Generation (SLVG)~\cite{qi2024signgen,saunders2020everybody}, and Pose to SL Video Generation (P2V)~\cite{shi2024pose}.

Among them, SLVG has recently gained increasing attention, aiming to generate realistic and expressive SL videos from spoken language inputs. Most existing studies adopt a two-stage generation paradigm, \ie first generating SL pose sequences (SLP) or other intermediate representations (\eg local regions, depth, or  optical  flow)~\cite{saunders2022signing,fang2023signdiff}, and then using these representations as conditions to synthesize corresponding SL frames (Pose2Video). Since pose sequences contain much less visual and spatial information, they are considerably easier to model than raw video frames, and thus are widely employed as intermediate representations. However, heavy dependence on intermediate modalities increases the complexity of model training, limits the scalability, and also introduces cumulative errors at each stage. In contrast, compared with pose sequences, video frames capture much richer spatial and appearance details, making the development of end-to-end\footnote{Here, end-to-end means that no intermediate modalities (pose, depth \etc) are produced or required during training or inference.} text-to-video SLVG models considerably more challenging.
 
Meanwhile, current mainstream text-to-video (T2V) frameworks in the general domain (\eg CogVideoX~\cite{yang2024cogvideox}, VideoCrafter~\cite{chen2024videocrafter2}, AnimateDiff~\cite{guo2023animatediff}, Wan~\cite{wan2025wan}) generally follow a de facto paradigm: a VAE is employed to compress videos into compact latent representations, enabling efficient modeling of high-dimensional video data. A DiT-like Transformer is then used to generate these latent video representations conditioned on text inputs, after which the pretrained VAE decoder reconstructs the final video frames. Other models like CogVideo~\cite{hong2022cogvideo}, VideoGPT~\cite{yan2021videogpt}, Phenaki~\cite{Villegas2022Phenaki} adopt a discrete token modeling paradigm, where videos are quantized into sequences of discrete token IDs using a VQ-VAE tokenizer. The generation process is then formulated as a sequence modeling task, in which autoregressive, discrete diffusion, or masked language modeling techniques are applied to predict the discrete video tokens. 
In this paper, we show that directly fine-tuning current state-of-the-art (SOTA) T2V models on the SLVG task may yield unsatisfactory results, since current T2V models directly adopt text sequences as conditions, which are not aligned with SL sequences and often fail to effectively guide SL video generation.

Different from previous works, we focus on the end-to-end Text-to-Video SLVG task and build a novel yet effective LLM-based Vector-Quantized Diffusion framework. We tackle two problems that are not addressed by existing  frameworks: \textit{(1) how to align the textual conditions with the SL sequence}, since SL exhibits different grammar and non-monotonic, variable-length alignments with spoken text, making it hard for textual conditions to directly guide SL video generation; and \textit{(2) how to ensure temporal coherence across consecutive frames/clips}, as natural sign language expressions require smooth and contextually consistent transitions between signs in SL. 
 
In our approach, LLMs act as powerful condition generators based on the input text. We decompose this challenging task into four essential components: (1) a Video VQ-VAE: SignVQ, which compresses and quantizes SL videos into discrete video tokens, enabling compact yet expressive latent representations;
(2) to address the misalignment between textual conditions and SL sequences, we introduce an \textbf{LLM-based conditioner} that generates semantically consistent conditions for each video clip, which is constrained by gloss sequences via a CTC-based alignment loss during training; (3) to ensure temporal coherence across frames, we propose a \textbf{Gaussian Condition Refiner} (GCR) that smooths the generated conditions over time to promote temporally coherent SL video generation; and (4) a VQ diffusion model is employed to generate latent video tokens conditioned on the refined condition sequence, which are then decoded by the SignVQ decoder to produce realistic sign language videos. 
During inference, the input text is first transformed into a sequence of semantically aligned conditions by the LLM conditioner. These conditions are refined by the GCR module and then passed into a VQ diffusion model to generate discrete video tokens, which are finally decoded into SL videos.  Our contributions are:
\vspace{-3mm}
\begin{itemize}  
\item We propose an LLM-based conditioner for end-to-end SLVG to generate fine-grained semantic conditions, incorporating a CTC alignment constraint to promote temporal semantic correspondence with the sign sequences.
\item  To enhance temporal coherence across consecutive frames, we introduce a Gaussian Condition Refiner that generates temporally smooth but semantically discriminative conditions, ensuring natural and consistent sign transitions in the generated videos.
\item To the best of our knowledge, we propose LLMCond, the first simple yet effective end-to-end T2V SLVG model that does not rely on auxiliary {visual} modalities (\eg pose, body parts, depth, or optical flow). Extensive experiments on public SL datasets show that our model achieves highly competitive performance {even compared with} two-stage models.
\end{itemize}

%% file: sec/2_related.tex
\section{Related Work}
\label{sec:related}
\paragraph{Text to Video Generation.} 
Current end-to-end T2V models generally fall into two paradigms: (1) modeling continuous latent representations extracted by VAE-based tokenizers, reformulating the video generation task as a problem of latent representation modeling;~\cite{yang2024cogvideox, chen2023videocrafter1, chen2024videocrafter2,guo2023animatediff}, or (2) compressing video into discrete token IDs via VQVAE-based tokenizers, converting the video generation task into a discrete token sequence modeling problem, which is then addressed using autoregressive, discrete diffusion or masked language modeling methods~\cite{hong2022cogvideo, yan2021videogpt, Villegas2022Phenaki}. The continuous-latent approach operates on low-dimensional real-valued embeddings, reducing pixel-level complexity and enabling latent diffusion pipelines to synthesize high-resolution frames. In contrast, the discrete-token paradigm compresses videos into sequences of quantized IDs and formulates video synthesis as a sequence modeling problem. Early influential works, such as VideoGPT~\cite{yan2021videogpt}, demonstrate that VQ-VAE tokenizers combined with autoregressive Transformers can effectively predict discrete video latents. Later models, including Phenaki~\cite{Villegas2022Phenaki}, extend this approach to variable-length T2V generation using discrete tokenizers and causal attention.
\textit{Despite these advances, in this paper, we show that directly fine-tuning current SOTA T2V models on SLVG task yields unsatisfactory results}, often producing inaccurate and temporally inconsistent hand movements, as they still struggle to (1) align textual conditions with sign sequences, since the ordering of signs often differs from that of spoken text, making it difficult for textual cues to effectively guide video generation, and (2) ensure temporal coherence across consecutive frames, which is essential for generating smooth and contextually consistent sign transitions. 

\paragraph{Sign Language Video Generation.}
{Sign language research is typically divided into understanding tasks (e.g., CSLR~\cite{wei2023improving}, SLT~\cite{gan2025mixsigngraph, shi2022open}) and generation tasks (SLP~\cite{saunders2020adversarial,saunders2021mixed,tang2025gloss,xie2024g2p,stoll2018sign,stoll2022there,yin2023gloss,rastgoo2021sign,liu2026bridging,zuo2025signs,tang2022gloss}, SLVG~\cite{saunders2022signing,fang2023signdiff,qi2024signgen,fang2025signllm}). The former converts SL sequences into gloss or spoken text, while the latter maps gloss or text back into sign language. Recently, SLVG has attracted increasing attention for its potential to synthesize realistic SL videos from text, improving information accessibility for the hearing impaired community.
Most existing SLVGs follow a two-stage design~\cite{saunders2022signing,fang2023signdiff}, where pose generation serves as an intermediate representation. Within this framework, autoregressive or diffusion models predict poses from text, while GAN or diffusion based models synthesize video frames from the generated pose sequences. Such dependence on intermediate modalities increases the complexity of SLVG model training, limits scalability, and can introduce cumulative errors at each stage. In contrast, image sequences contain richer visual details, making end-to-end SLVG modeling substantially more challenging. Early attempts, such as SignGen~\cite{qi2024signgen}, employ auxiliary conditions (\eg optical flow, pose, depth) during training; however, these signals are unavailable during inference, which highly constrains the generation quality. In this work, we propose a simple yet effective end-to-end SLVG model that eliminates reliance on auxiliary visual conditions in training and inference.

%% file: sec/3_method.tex
\section{Method}

\begin{figure}[t]
  \centering
    \includegraphics[width=1.0\columnwidth]{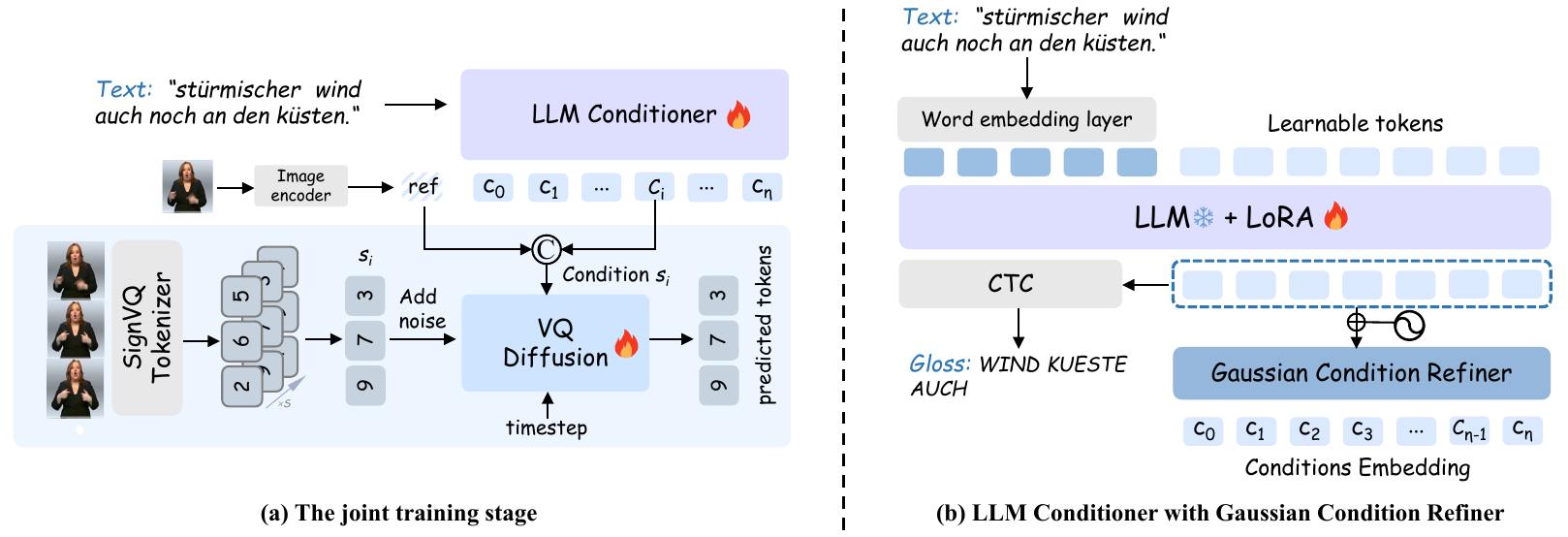}
    \vspace{-1mm}
    \caption{Overview of LLMCond. (a) Joint training, where SignVQ encodes videos into discrete tokens and VQ-Diffusion generates video tokens conditioned on a reference image and conditions. (b) LLM conditioner, which fuses text embeddings with learnable tokens, enforces gloss alignment via CTC, and refines conditions using a Gaussian refiner for temporal consistency.}
   \label{fig:framework} 
  \vspace{-4mm}
\end{figure}

\paragraph{Overall Framework.}  
We formulate the SLVG task as a condition-based SL clip generation process which consists of five components: video quantization, condition generation, condition refinement, video latent generation, and video latent dequantization. Specifically, given a text sequence $x=\{x_i\}_{i=0}^ { \varsigma }$ with $\varsigma$ words, the goal of SLVG is to generate a  SL video with $F$ frames $f=\{f_i\}_{i=0}^{F}$ based on input $x$.  As shown in Figure~\ref{fig:framework},  our model tackles the SLVG as follows: (1) a video tokenizer that compresses and quantizes video sequences into discrete tokens $v=\mathcal{VQ}_E({f})$, where $v = \{v_i\}_{i=1}^{\eta}$ and $\eta$ denotes the temporal length after the tokenizer's downsampling; (2) an LLM-based conditioner that generates temporally aligned condition embeddings $c = \mathcal{LLM}(x)$, where $\mathbf{c}=\{\mathbf{c}_i\}_{i=0}^\eta$; (3) a  Gaussian Condition Refiner (GCR) that smooths the generated conditions over time to alleviate jitters in generated videos; (4) a VQ diffusion that produces discrete video tokens  $\hat{v} = \mathcal{LDM}(\mathbf{c}, c_{\text{ref}})$, where $c_{\text{ref}}$ is the reference image embedding used to guide the generation; and (5) the predicted tokens are subsequently decoded into full videos by the video tokenizer decoder $\hat{f} =\mathcal{VQ}_D(\{\hat{v}_i\}_{i=0}^{\eta})$. 
 

\subsection{Sign Language Video Tokenizer} 
Current video generation models typically use VAE or VQ-VAE to compress a video $f$ into a latent representation, reducing video generation to latent modeling. Inspired by VQ-VAE-2~\cite{razavi2019generating}, we propose SignVQ, a sign language tokenizer that encodes raw SL videos into discrete tokens with a two-level codebook.
Given a video $f \in \mathbb{R}^{F \times H \times W \times 3}$, where $F$, $H$, and $W$ denote the number of frames, height, and width, the encoder $\mathcal{VQ}_E$ first performs spatial and temporal downsampling by factors of $d_b$ and $d_f$ via 3D convolutional blocks, producing a coarse spatiotemporal feature map $e_b$. This representation is further spatially downsampled to obtain a high-level feature $e_t$.
\begin{equation}
e_t, e_b = \mathcal{VQ}_E(f).
\end{equation}
Next, two vector quantizers, $Q_t$ and $Q_b$, quantize the top and bottom features $e_t$ and $e_b$, respectively:
\begin{equation}
v_t = Q_t(e_t), \quad
v_b = Q_b([U_t(\hat{e}_t), e_b]),
\end{equation}
where $U_t$ denotes the upsampling operation applied to the quantized top-level features $\hat{e}_t$. The resulting discrete indices $v_t$ and $v_b$ form a sequence of quantized video tokens $v = \{v_i\}_{i=1}^{\eta}$, where $v_i = \{v_{t,i}, v_{b,i}\}$. The total video id length \ie (the number of ids) is computed as:
\begin{equation} 
\small
L_v = \eta \times (L_{v_t} + L_{v_b}) = \frac{F}{d_f} \times \left( \frac{H \times W}{d_t^2} + \frac{H \times W}{d_b^2} \right).
\end{equation}
Thus, the SLVG objective shifts from modeling pixel-space likelihood $p(f|x)$ to discrete latent likelihood $p(v|c)$.  
During reconstruction, the SignVQ decoder $\mathcal{VQ}_D$ retrieves the nearest embeddings from the corresponding codebooks using discrete indices $\{v_i\}_{i=1}^{\eta}$ and reconstructs the SL video $\hat{f}$ as:
\begin{equation}
\hat{f} = \mathcal{VQ}_D(v_t, v_b),
\label{equ:vqdecoder}
\end{equation}

\subsection{LLM Conditioner}
Traditionally, T2V models directly embed text inputs as condition embeddings, using pretrained language models such as CLIP or BERT. While effective in general domains, this approach often fails in SLVG, due to the unique grammatical structure of SL, \ie spoken texts are not monotonically aligned with sign sequences. Consequently, these embeddings 
fail to capture semantic alignment with SL sequences, leading to temporally inconsistent and semantically distorted SL video generation.  

To bridge the gap between spoken text and SL sequence, we design an LLM-based conditioner that provides temporally and semantically aligned conditional embeddings, to guide SL frame generation. As shown in Figure~\ref{fig:framework}(b), given a text sequence $x$, and its corresponding embedding $e_x =\{e_x^i\}_{i=0}^{\ell}$, we introduce a sequence of learnable condition tokens
$e_l=\{e_l^i\}_{i=0}^{\eta }$ that serve as alignment anchors between text and sign language. The text embeddings and learnable condition tokens are then concatenated and fed into the LLM for contextual modeling:  
\begin{equation} 
     h =\mathcal{LLM}([e_x, e_l]) 
\end{equation} 
We take the hidden states $h$ at learnable tokens as the target conditional embeddings, denoted as $c=h[\ell:,]$. To ensure that LLM-generated conditions preserve text semantics and temporal consistency with SL videos, we impose a CTC alignment loss $\mathcal{L}_{CTC}$ between the generated conditions and corresponding gloss sequences $g$ during training.
\begin{equation}
\label{equ:ctcloss}
\Theta_{\text{LLM}}^{*},\, e_l^{*} 
 \leftarrow 
 \mathop{\arg\min}_{\Theta_{\text{LLM}},\, e_l} 
\mathbb{E}_{(x, g) \sim \mathcal{D}}
\big[
\mathcal{L}_{\text{CTC}}(c * W_g, g)
\big],
\end{equation}
where $\Theta_{\text{LLM}}$ denotes the trainable parameters of LLM and $W_g$ denotes the classification layer weights. Through these constraints, the condition embeddings $c$ are temporally aligned with the gloss sequence, allowing each embedding to independently provide semantically aligned and temporally consistent guidance for SL video generation. \textit{The length of the learnable token $\ell$ is determined by a length prediction module; see the Appendix for details.}

\begin{wrapfigure}{r}{0.65\textwidth} 
\centering 
    \includegraphics[width=0.65\columnwidth]{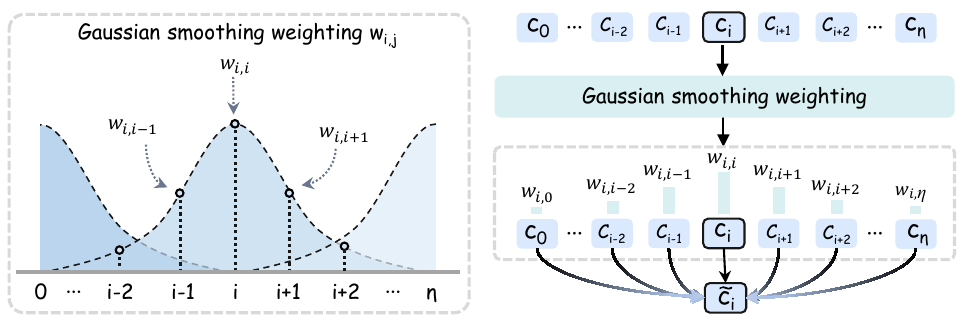}
    \vspace{-4mm}
   \caption{The Gaussian Condition Refiner (GCR) module.} 
   \label{fig:gaussian}  
  \vspace{-5mm}
\end{wrapfigure}

\subsection{Gaussian Condition Refiner} 
Although the LLM conditioner module ensures that the generated conditions 
$c$ are temporally aligned with the SL sequences, aligning conditions $c$  with the gloss sequence discretizes its distribution, which may lead to discontinuities between adjacent sign clips. To alleviate this issue, we further smooth $c$ using a Gaussian Condition Refiner (GCR) to promote seamless transitions across clips.

Specifically, as shown in Figure~\ref{fig:gaussian}, given a condition sequence $c=\{c_i\}_{i=0}^{\eta}$, we compute Gaussian weights $\{w_{i,j}\}_{j=0}^{\eta}$ centered at each position $i$ and apply them to neighboring conditions via weighted summation, to obtain a refined condition $\tilde c_i$. This encourages smooth transitions and temporally coherent conditioning signals, where $\sigma$ controls the smoothing strength. 
\begin{equation} 
\small
\tilde{c}_i =
\frac{\sum_{j=0}^{\eta } \left \{ \exp\!  \left(-\tfrac{(j-i)^2}{2\sigma^2}\right)\cdot {c}_j \right \} }
{\sum_{j=0}^{\eta } \exp\!\left(-\tfrac{(j-i)^2}{2\sigma^2}\right)},
\end{equation}  

\subsection{VQ-Diffusion Model for SLVG}
After obtaining the refined condition sequence $c=\{c_i\}_{i=0}^{\eta}$, we adopt a VQ-Diffusion model~\cite{gu2022vector} to generate $v_i$ conditioned on $c_i$. 
During training, VQ-Diffusion gradually corrupts $v_i$ via a fixed Markov chain $q(v_{i,t}|v_{i,t-1})$, \eg randomly replacing tokens of $v_{i,t-1}$, where $v_{i,0}=v_i$. 
By the Markov property, we derive the probability of $v_{i,t}$ as:
\begin{equation}
\small
q(v_{i,t} | v_{i,0}) = e^\top(v_t)\, \overline{Q}_t\, e(v_0), \quad   \overline{Q}_t = Q_t \cdots Q_1,
\label{eq:forward_diffusion}
\end{equation}
where $e(v_t) \in \mathbb{R}^{K+1}$ is the one-hot representation of a token id (with an additional id for the \texttt{[MASK]} token) and $K$ is codebook size. The transition matrix $Q_t \in \mathbb{R}^{(K+1) \times (K+1)}$ is defined as:
\begin{equation}
\tiny
Q_t =
\left[
\begin{array}{cccccc}
\alpha_t + \beta_t & \beta_t & \cdots & 0 & \cdots & 0 \\
\beta_t & \alpha_t + \beta_t & \cdots & \vdots & \cdots & 0 \\
\vdots & \vdots & \ddots & \vdots & & \vdots \\
\beta_t & \cdots & \cdots & \alpha_t + \beta_t & \cdots & 0 \\
\gamma_t & \gamma_t & \cdots & \gamma_t & \cdots & 1 \\
\end{array}
\right].
\end{equation}
After a fixed number of $T$ timesteps, the $v_{i, T}$ becomes pure noise tokens. Starting from the noise $v_{i, T}$, the reverse process gradually denoises the latent variables, by sampling from the reverse distribution $q(v_{i, t-1 }|v_{i, t}, \tilde{c}_i, c_{ref} )$, where $c_{ref}$ is reference image embedding, and then restores the $v_{i,0}$.  
For reverse process, we train a Transformer denoiser to model the probability $p_\theta(v_{i, t-1}\!\mid v_{i,t},\tilde c_i, c_{ref})$ as:
\begin{equation}
\small
p_\theta(v_{i,t-1} | v_{i,t}, \tilde c_i, c_{ref})
= \sum_{\tilde v_{i,0}} q(v_{i,t-1} | v_{i,t}, \tilde v_{i,0})\,
p_\theta(\tilde v_{i,0} | v_{i,t}, \tilde c_i, c_{ref}),
\label{eq:reverse_aggregate}
\end{equation}
Then, the VQ-Diffusion module is trained under:
\begin{equation} 
\small
\begin{aligned}
\label{equ:vqdiffkl}
\mathcal{L}_{\mathcal{DI}}
&=
\mathbb{E}_{t \sim \mathcal{U}(1, T),\, ({v}_{i,0}, v_{i,t}) \sim q}
\Big[
D_{\mathrm{KL}}\!\big(
q(v_{i,t-1} \mid v_{i,t}, {v}_{i,0}) \;\|\;
p_{\theta}(v_{i,t-1} \mid v_{i,t}, \tilde{c}_i, c_{ref})
\big)
\Big].
\end{aligned} 
\end{equation}  
\subsection{Training and Inference}
\paragraph{Training.}The final training process of LLMCond first trains SignVQ model with reconstruction loss, embedding loss and commitment loss as follows: 
 \begin{equation}
\small
\mathop{\arg\min}_{\Theta_{\mathcal{VQ}}} \mathbb{E}_{(f) \sim \mathcal{D}}
\Big[
\| f - \hat{f} \|_2^2 
+ \| \mathrm{sg}[\hat{e}] - e \|_2^2 
+ \beta \| \hat{e} - \mathrm{sg}[e] \|_2^2
\Big],
\end{equation} where $||\cdot||_{2}$ denotes the L2 norm and $\beta$ is a weighting coefficient.
After the SignVQ is trained, we train the {LLM Conditioner} together with the VQ-Diffusion model using the following joint loss: 
\begin{equation}
\begin{aligned}
\label{equ:totalloss}
\small 
 \mathop{\arg\min}_{\Theta_{\mathcal{LLM}}, \Theta_{\mathcal{DI}}} 
&\mathbb{E}_{(f, x, g) \sim \mathcal{D}}
\Big[
\alpha \, \mathcal{L}_{\text{CTC}}(c * W_g, g) + \lambda \sum_{i=0}^{\eta} \mathcal{L}_{DI}(v_i, \tilde{c}_i, c_{ref})
\Big],
\end{aligned} 
\end{equation} 
where $\alpha$ and $\lambda$ are hyperparameters. The condition sequence $c$ is optimized under CTC supervision and diffusion constraints, encouraging temporally aligned, visually coherent sign video generation.

\paragraph{Inference.} During inference, the text sequence is fed into the LLM conditioner to generate the condition sequence. The VQ-diffusion model then produces a sequence of latent discrete video tokens $\{\hat{v}_i\}_{i=0}^{\eta}$ based on $\tilde{c}$, which are decoded by the SignVQ decoder into the final SL video $\hat{f}$. {No gloss annotations or intermediate modalities are required or generated in this process.}

%% file: sec/4_experiment.tex
\section{Experiments}
\label{sec:experiment}

\paragraph{Datasets.}   We evaluate our model on two widely used SL datasets with both gloss and text annotations: (1) \textit{Phoenix14T}~\cite{camgoz2018neural}, a German SL corpus from weather broadcasts, consists of 8257 videos from nine signers with 1066 glosses and 2877 German words, split into 7096, 519, and 642 videos for training, validation, and testing, respectively. (2) \textit{CSL-Daily}~\cite{zhou2021improving}, a Chinese SL dataset with a vocabulary of 2000 glosses and 2343 Chinese words, comprising 18401, 1077, 1176 videos for training, validation and testing respectively.

\paragraph{Evaluation Metrics.} To evaluate \textbf{semantic accuracy}, we employ a pretrained SLT model~\cite{gan2025mixsigngraph} to translate generated videos back into text. The resulting translations are compared with reference sentences using BLEU~\cite{papineni2002bleu} and ROUGE-L~\cite{lin2004automatic}. These scores reflect how accurately generated videos convey sign meaning. To evaluate \textbf{video quality}, we adopt common T2V metrics, including FID~\cite{heusel2017gans}, PSNR, SSIM, and LPIPS for reconstruction, and FVD~\cite{unterthiner2018towards} for generation.
These metrics measure pixel-level similarity, perceptual fidelity, and temporal consistency between generated and GT videos.
 
\vspace{-2mm}
\paragraph{Implementation Details.}
(1) \textit{SignVQ}: In the baseline setup, the spatial and temporal downsampling factors $d_f$, $d_b$, and $d_t$ are set to 4, 16, and 32, respectively. The latent channel dimension is 128.
(2) \textit{LLM Conditioner}: We employ the LLaMA-3.2 1B model fine-tuned using LoRA~\cite{hu2022lora}. The LoRA modules are inserted into the $\texttt{q\_proj}$, $\texttt{v\_proj}$, and $\texttt{o\_proj}$ layers, with the rank set to 64. 
(3) \textit{GCR}: The Gaussian smoothing parameter $\sigma$ is set to 0.5.
(4) \textit{VQ Diffusion}: We adopt a VQ-Diffusion backbone consisting of 15 Transformer layers. The generated condition tokens $\tilde{c}$ are fused through cross-attention layers in every Transformer block, while reference image conditions are fused at the 3rd, 9th, and 12th layers. \textit{More implementation details are provided in the Appendix.}

%% file: sec/4.1_compare.tex
\section{Comparisons}

\input{table/comparsion}

\input{table/compare_CSL}

We use MixSignGraph~\cite{gan2025mixsigngraph}, an SLT model, to evaluate videos generated by our SLVG model and the baseline methods. For a fair comparison, we reproduce several open-source baselines and report their results under a unified evaluation protocol. We compare our method with end-to-end SLVG models as well as two-stage approaches that rely on explicit auxiliary modalities, such as pose or depth, during training or inference. Results are reported in Tables~\ref{tab:SLVGphoenix14t} and~\ref{tab:SLVGcsl}.

\paragraph{Evaluation of Generation on Phoenix14T.}
We compare two groups of baselines on Phoenix14T, reporting back-translation results and generated-video FVD (gFVD).
\textit{(1) Comparison with End-to-End T2V Models.}
We fine-tune three representative open-source T2V models on our dataset, including continuous latent-based CogVideoX-2B and Wan2.1-1.3B, and discrete latent-based VideoGPT.
As shown in Table~\ref{tab:SLVGphoenix14t}, directly adapting general-purpose T2V models to SLVG leads to inaccurate sign generation and unsatisfactory performance.
These models are mainly optimized for globally coherent visual content and smooth motion, rather than fine-grained human actions, making it difficult to synthesize subtle and temporally precise sign gestures.
This limitation is also noted in the official documentation~\footnote{\url{https://huggingface.co/docs/diffusers/training/cogvideox}}, which reports strong global consistency and appearance modeling but limited capability for detailed human actions.
For SignGen, although it uses pose, depth, and optical flow during training, its reliance on these modalities makes it difficult to generate semantically correct sign videos when they are unavailable during inference.
\textit{(2) Comparison with Two-Stage SLVG Models.}
We further compare with current two-stage SLVG models.
Without requiring intermediate modalities during training or inference, our model achieves SOTA back-translation performance, demonstrating the effectiveness of the LLM-based conditioner in aligning text conditions with sign sequences.
Moreover, it obtains lower gFVD compared to SignGAN, which uses pose as an intermediate representation.

\paragraph{Evaluation of Generation on CSL-Daily.}
We further evaluate our model on the CSL-Daily, comparing it with both T2V models (Wan2.1 \etc) and SLVG models (SignGAN \etc). As shown in Table~\ref{tab:SLVGcsl}, our method achieves clear improvements in both back-translation accuracy and video quality.
Compared with the T2V model Wan2.1, ROUGE increases from 12.33 to 24.80, while gFVD decreases from 1985.47 to 1412.48.
Compared with the SLVG model SignGAN, our LLMCond achieves gains of +11.38 ROUGE, along with a 533.79 reduction in gFVD.

\begin{figure*}[t]
  \centering
    \includegraphics[width=1.0\linewidth]{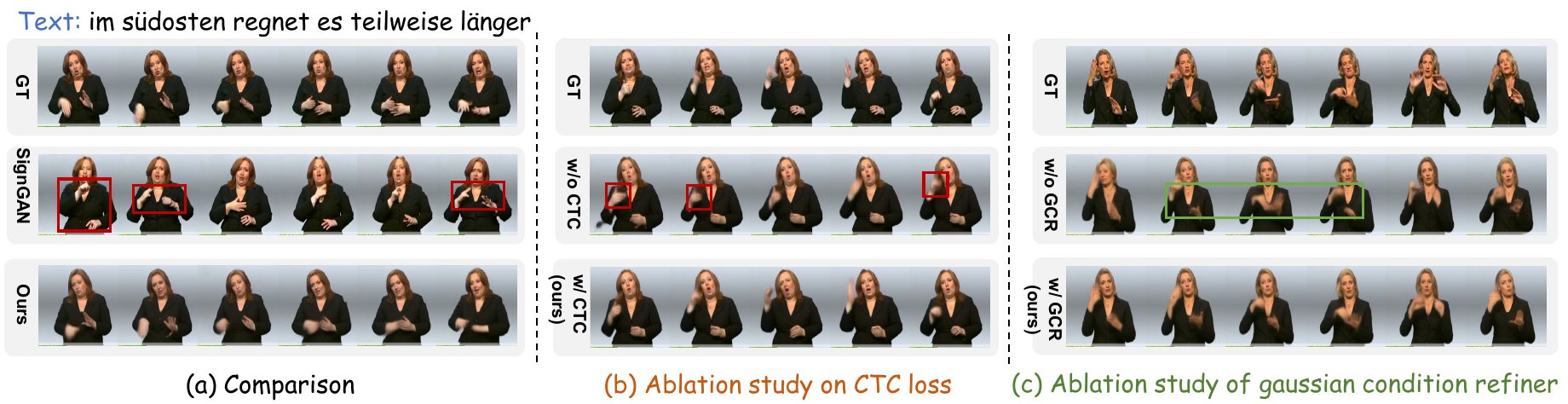}
    \vspace{-2mm}
   \caption{(a) Qualitative comparison on Phoenix14T with SignGAN.  (b) Ablation study on CTC loss. (c) Ablation study on Gaussian Condition Refiner (GCR). Red boxes denote incorrect motions, while green boxes indicate temporal discontinuities.}
   \label{fig:compare} 
  \vspace{-4mm}
\end{figure*}

\section{Qualitative Results} 
\paragraph{Visualization Comparison.}
To intuitively demonstrate the generation quality of our model, we visualize one sample from the Phoenix14T dataset and compare it with the previous SLVG model SignGAN. As shown in Figure~\ref{fig:compare} (a), compared with the ground-truth SL sequences (GT), our model produces an SL sequence that more closely matches the ground truth in both visual quality and temporal coherence. In contrast, SignGAN often exhibits misaligned gestures, leading to reduced semantic accuracy. These results indicate that LLMCond models spatiotemporal dependencies more effectively, achieving tighter alignment between text semantics and sign sequences.

\paragraph{Visualization Ablation.}
As shown in Figure~\ref{fig:compare} (b), we compare the ground truth (GT), a variant trained without the CTC constraint (w/o CTC), and our model (w/ CTC).
In Figure~\ref{fig:compare} (c), we further compare GT, a variant without Gaussian condition refiner (w/o GCR), and our model (w/ GCR).
It can be clearly observed that removing the CTC constraint and GCR respectively results in semantic inaccuracy and temporal inconsistency, highlighting the importance and effectiveness of our proposed design. \textit{Please refer to the supplementary materials for more visualization.}

%% file: table/comparsion.tex
\begin{table}[t]
\centering
\caption{Comparison of SLVG performance on Phoenix14T. (P: pose, H: 3D hands, D: depth, O: optical flow, * denotes our finetuning results. The same notation applies to the tables below.)}
\resizebox{1.0\textwidth}{!}{
\begin{tabular}{l|cccc|cccccc|cccccc}
\multirow{2}{*}{Two-Stage}
& \multicolumn{4}{c|}{Extra Data}
& \multicolumn{6}{c|}{DEV}
& \multicolumn{6}{c}{TEST} \\
\cline{2-17}
& P & H & D & O
& ROUGE & BLEU1 & BLEU2 & BLEU3 & BLEU4 & gFVD$\downarrow$
& ROUGE & BLEU1 & BLEU2 & BLEU3 & BLEU4 & gFVD$\downarrow$ \\
\midrule
ProTran~\cite{saunders2020progressive}~+~Ani~\cite{hu2024animate}    & \checkmark &  &  &  & 9.74 & 9.09 & 2.95 & 1.51 & 0.75 & 2125.19 & 10.33 & 9.79 & 3.60 & 1.93 & 0.82 & 2017.90 \\
SignGAN~\cite{saunders2022signing}     & \checkmark &  &  &  & 13.90 & 13.27 & 4.91 & 2.55 & 1.36 & 1986.33 & 15.97 & 14.02 & 6.74 & 3.86 & 2.67 & 1918.42 \\
\midrule
\multicolumn{17}{l}{\textit{End-to-End}} \\
\midrule
SignGen~\cite{qi2024signgen}     & \checkmark &  & \checkmark & \checkmark & 10.41 & 9.36 & 2.15 & 0.83 & 0.39 & 2565.03 & 10.08 & 10.61 & 2.47 & 1.03 & 0.45 & 2553.14 \\
VideoGPT*~\cite{yan2021videogpt}   &  &  &  &  & 10.83 & 9.85 & 3.60 & 1.71 & 0.86 & 2021.74 & 11.10 & 10.71 & 3.66 & 2.03 & 0.87 & 2011.89\\
CogVideoX*~\cite{yang2024cogvideox}   &  &  &  &  & 11.24 & 11.18 & 3.75 & 2.06 & 1.19 & 2008.25 & 11.67 & 11.32 & 4.18 & 2.02 & 1.14 & 1998.81\\
Wan2.1*~\cite{wan2025wan}  &  &  &  &  & 12.14 & 11.75 & 4.36 & 2.48 & 1.26 & 2074.31 & 12.56 & 12.79 & 5.64 & 2.41 & 1.40 & 1894.75\\
\baseline{\textbf{LLMCond}}& \baseline{}  & \baseline{} & \baseline{} & \baseline{} & \baseline{\textbf{26.57}} & \baseline{\textbf{27.06}} & \baseline{\textbf{14.95}} & \baseline{\textbf{10.68}} & \baseline{\textbf{7.93}} & \textbf{\baseline{1124.10}} & \baseline{\textbf{27.29}} & \baseline{\textbf{26.55}} & \baseline{\textbf{15.36}} & \baseline{\textbf{10.43}} & \baseline{\textbf{7.91}} & \textbf{\baseline{1107.46}}  \\
\end{tabular}
}  
\vspace{-3mm}
\label{tab:SLVGphoenix14t}
\end{table}

%% file: table/compare_CSL.tex
\begin{table*}[t]
\centering
\caption{Comparison of SLVG performance on the CSL-Daily dataset.}
\vspace{-2mm}
\resizebox{1.0\textwidth}{!}{
\begin{tabular}{l|ccc|cccccc|cccccc}
\multirow{2}{*}{Two-Stage}
& \multicolumn{3}{c|}{Extra Data}
& \multicolumn{6}{c|}{DEV}
& \multicolumn{6}{c}{TEST} \\
\cline{2-16}
& P & D & O
& ROUGE & BLEU1 & BLEU2 & BLEU3 & BLEU4 & gFVD$\downarrow$
& ROUGE & BLEU1 & BLEU2 & BLEU3 & BLEU4 & gFVD$\downarrow$ \\
\midrule
ProTran~\cite{saunders2020progressive}~+~Ani~\cite{hu2024animate}   & \checkmark &  &  & 8.93 & 8.81 & 2.74 & 1.55 & 0.63 & 2158.04 & 9.41 & 9.48 & 3.50 & 1.82 & 0.69 & 2077.45 \\
SignGAN~\cite{saunders2022signing}     & \checkmark &  &  & 14.02 & 14.74 & 5.17 & 2.68 & 1.73 & 1923.53 & 13.42 & 13.85 & 4.11 & 1.86 & 1.09 & 1946.27 \\ 
\midrule
\multicolumn{16}{l}{\textit{End-to-End}} \\
\midrule
SignGen~\cite{qi2024signgen}     & \checkmark & \checkmark & \checkmark & 10.44 & 10.98 & 3.04 & 2.27 & 1.30 & 2406.46 & 11.45 & 11.39 & 4.24 & 2.76 & 1.38 & 2551.87 \\
VideoGPT*~\cite{yan2021videogpt}   &  &  &  & 9.17 & 9.55 & 2.52 & 1.73 & 0.69 & 2674.20 & 9.60 & 9.91 & 3.16 & 1.49 & 0.66 & 2694.93 \\
CogVideoX*~\cite{yang2024cogvideox}   &  &  &  & 11.56 & 10.22 & 3.02 & 2.18 & 1.66 & 1944.53 & 11.42 & 10.90 & 4.21 & 2.94 & 1.28 & 1962.08\\
Wan2.1*~\cite{wan2025wan}  &  &  &  & 11.87 & 11.20 & 3.39 & 2.36 & 1.74 & 1932.50 & 12.33 & 12.61 & 4.59 & 3.25 & 1.58 & 1985.47\\
\baseline{\textbf{LLMCond}}& \baseline{}  & \baseline{} & \baseline{} &\baseline{\textbf{24.59}} & \baseline{\textbf{24.31}} & \baseline{\textbf{14.93}} & \baseline{\textbf{9.65}} & \baseline{\textbf{7.78}} & \baseline{\textbf{1500.27}} & \baseline{\textbf{24.80}} & \baseline{\textbf{25.26}} & \baseline{\textbf{15.04}} & \baseline{\textbf{9.89}} & \baseline{\textbf{7.94}} & \baseline{\textbf{1412.48}}  \\
\end{tabular}}
\vspace{-3mm}
\label{tab:SLVGcsl}
\end{table*}

%% file: sec/5_performance.tex
\section{Ablation Study}
In this subsection, we report ablations on the Phoenix14T dataset that verify the effectiveness of LLMCond. \textit{Please refer to the Appendix for more ablation study.}\vspace{-3mm}

\paragraph{Reconstruction Results of Different Compression Rate.} 
We first analyze how compression affects quantization and reconstruction in SignVQ.
As reported in Table~\ref{tab:Revae}, higher compression degrades VQ-VAE reconstruction but shortens the latent sequence. Increasing compression from $2\times8\times8$ to $4\times16\times16$ (time$\times$height$\times$width) raises FVD, while making latent codes $8\times$ shorter. Overall, stronger compression trades reconstruction fidelity for compactness. We also evaluate key local regions (\eg hands and face) cropped using HRNet keypoints, where the model captures fine-grained SL motion details. To balance compression and reconstruction, we adopt $4\times16\times16$ compression.


\paragraph{Ablation study on Conditioner Modules.}
We investigate the impact of the LLM Conditioner and GCR in Table~\ref{tab:ablation_conditioner}.
As shown, introducing the LLM Conditioner with CTC supervision significantly improves semantic accuracy and video quality, demonstrating its effectiveness in aligning textual and visual representations.
Removing CTC supervision causes a severe performance drop, indicating that temporal alignment is essential for guiding the conditioner.
Similarly, excluding GCR reduces generation quality, as indicated by higher gFVD, suggesting that Gaussian smoothing helps refine temporal coherence.
Overall, our model with both the LLM Conditioner and GCR achieves the best performance across all metrics, confirming the effectiveness of the proposed components.

\input{table/videoVQVAE}
\input{table/Ab_LLMCon}

\begin{wrapfigure}{r}{0.55\textwidth} 
\centering 
\vspace{-4mm} 
    \includegraphics[width=0.55\columnwidth]{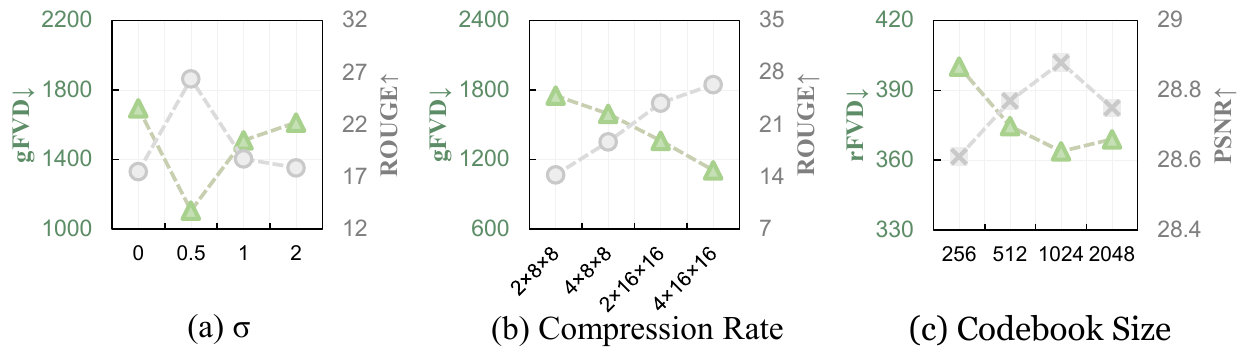}
   \caption{Effect of SignVQ and GCR hyperparameters. Triangles: gFVD; circles: ROUGE; crosses: PSNR.}
      \label{fig:abacodebook} 
  \vspace{-5mm}
\end{wrapfigure}

\paragraph{Ablation Study on $\sigma$ in Gaussian Smoothing.}
To investigate the impact of Gaussian standard deviation $\sigma$ in the GCR module, we conduct experiments with different $\sigma$ values. A larger $\sigma$ strengthens the influence of neighboring conditions on each clip's generation. As shown in Figure~\ref{fig:abacodebook}(a), the generated results vary significantly with $\sigma$, and the model performs best when $\sigma = 0.5$. A smaller $\sigma$ makes each clip more independent, causing discontinuous motion, whereas an excessively large $\sigma$ leads to over-smoothing with overly similar neighboring clips.

\paragraph{Generation Results of Different Compression Rate.}
We analyze the effect of video compression on LLMCond. As shown in Figure~\ref{fig:abacodebook}(b), stronger compression improves ROUGE and FVD, despite modestly reducing VQ reconstruction quality (Table~\ref{tab:Revae}). The shorter latent sequence eases generation, outweighing reconstruction loss and improving overall performance.



\paragraph{Ablation Study on Codebook Size for Quantization.}
We evaluate tokenizer codebook size for 3D video reconstruction in Figure~\ref{fig:abacodebook}(c). Small codebooks (256) increase reconstruction error, lowering PSNR and raising FVD. Performance improves up to 1024, where PSNR peaks and FVD is lowest. Increasing to 2048 gives no gain and slightly degrades generation.

\paragraph{Ablation Study on signer-independent test.} 
To evaluate generalization to unseen signers, we re-split Phoenix14T,
using one signer as the test set and the remaining eight for training.
As shown in Table~\ref{tab:SignerInde}, performance only slightly decreases
compared to the Norm-Split (ROUGE 27.29 to 25.67), indicating that LLMCond achieves comparable performance to the standard split, with ROUGE-L decreasing only slightly from 27.29 to 25.67, suggesting good generalization to unseen signers.


\paragraph{Ablation Study on Generation Models.}
To examine the effectiveness of our VQ-Diffusion model, we compare it with several representative discrete sequence generation approaches, including an AR-based model (Llama3.2-1B), a masked diffusion model (MDM)-based variant (LLaDA), and VQ-Diffusion. As shown in Table~\ref{tab:GenerationType}, the AR-based Llama model yields the weakest performance, as it struggles to capture spatial dependencies in sign sequences. In contrast, our VQ-Diffusion model achieves the best results in both semantic accuracy and video quality, demonstrating its superior capability for coherent SL video generation.

\input{table/Ab_VQvsAR}

\paragraph{Comparison of Different Representation Paradigms.}
In mainstream text-to-video frameworks, videos are often represented either as continuous or discrete latent representations. Here, we evaluate our framework under different representation paradigms. We replace the VQ-VAE with a VAE whose latent channel is set to 16, and replace the discrete diffusion model with a continuous diffusion model, while keeping other components unchanged. As shown in Table~\ref{tab:gen_paradigms}, models based on continuous representations also achieve good performance, demonstrating the generalizability of our method. However, in our experiments, discrete-based models achieve slightly better generation results on both back-translation metrics and video quality.



\paragraph{Ablation of Training Strategies.}
\input{table/model_para}
We also present generation results under different training strategies. In the first strategy, the LLM Conditioner is trained with a CTC constraint and then frozen before training the VQ-Diffusion model (denoted as `Separate'). In the second strategy, the LLM Conditioner and VQ-Diffusion model are trained jointly under the constraint of Equation~13 (denoted as `Joint'). As shown in Table~\ref{tab:training}, our joint training strategy achieves better performance.

%% file: table/Ab_LLMCon.tex
\begin{figure*}[t]
\begin{minipage}[t]{0.50\linewidth}
    \centering
    \setlength{\tabcolsep}{4pt}
    \captionof{table}{Effect of different variants of SignVQ.}
    \label{tab:Revae}
    \renewcommand{\arraystretch}{1.15}
    \resizebox{\linewidth}{!}{
    \begin{tabular}{c|cccc}
    \toprule
        Variants & Baseline & A & B & C \\ 
        \midrule
        Compression & 4$\times$16$\times$16 & 2$\times$16$\times$16 & 4$\times$8$\times$8 & 2$\times$8$\times$8 \\ 
        Latent Codes & $\frac{L}{4}$$\times$245 & $\frac{L}{2}$$\times$245 & $\frac{L}{4}$$\times$980 & $\frac{L}{2}$$\times$980 \\ 
        Codebook Size & 2048 & 1024 & 1024 & 512 \\  
        rFID$\downarrow$ & 0.23 & 0.16 & 0.13 & 0.03 \\ 
        rLPIPS$\downarrow$ & 0.0611 & 0.0545 & 0.0492 & 0.0425 \\ 
        rFVD$\downarrow$ & 363.75 & 319.60 & 259.27 & 194.46 \\ 
        rPSNR$\uparrow$ & 28.88 & 29.86 & 30.68 & 32.01 \\
        rSSIM$\uparrow$ & 0.89 & 0.90 & 0.91 & 0.93 \\ 
        Hand SSIM$\uparrow$ & 0.89 & 0.90 & 0.91 & 0.92 \\ 
        Face SSIM$\uparrow$ & 0.88 & 0.89 & 0.90 & 0.91 \\ 
    \bottomrule
    \end{tabular}}
\end{minipage}
\hfill
\begin{minipage}[t]{0.47\linewidth}
    \centering

    \setlength{\tabcolsep}{4pt}
    \captionof{table}{Effect of LLM conditioner and GCR.}
    \label{tab:ablation_conditioner}
    \renewcommand{\arraystretch}{1.2}
    \resizebox{0.99\linewidth}{!}{
    \begin{tabular}{cc|cccc}
    \toprule 
    CTC & GCR & ROUGE-L & BLEU-1 & BLEU-4 & gFVD$\downarrow$ \\
    \midrule
          &            & 20.95 & 21.22 & 5.20 & 1533.16 \\
    \checkmark &       & 24.41 & 24.92 & 6.98 & 1257.95 \\
          & \checkmark & 21.36 & 21.60 & 5.53 & 1412.28 \\
    \baseline{\checkmark} & \baseline{\checkmark} 
    & \baseline{\textbf{27.29}} 
    & \baseline{\textbf{26.55}} 
    & \baseline{\textbf{7.91}} 
    & \baseline{\textbf{1107.46}} \\
    \bottomrule
    \end{tabular}}

    \vspace{2mm}

    \setlength{\tabcolsep}{4pt}
    \captionof{table}{Ablation on Signer-Independent test.}
    \label{tab:SignerInde}
    \renewcommand{\arraystretch}{1.15}
    \resizebox{0.95\linewidth}{!}{
    \begin{tabular}{l|cccc}
    \toprule
    Split & ROUGE-L & BLEU-1 & BLEU-4 & gFVD$\downarrow$ \\
    \midrule
    Norm-Split & 27.29 & 26.55 & 7.91 & 1107.46 \\ 
    Signer-Independent & 25.67 & 25.41 & 7.69 & 1345.36 \\ 
    \bottomrule
    \end{tabular}}

\end{minipage}
\vspace{-5mm}
\end{figure*}

%% file: table/Ab_VQvsAR.tex
\begin{figure*}[t]
\begin{minipage}[t]{0.5\linewidth}
     \centering
  \setlength{\tabcolsep}{4pt}
     \captionof{table}{\label{tab:GenerationType}Comparison of different generation models.}
  \setlength{\tabcolsep}{4pt}
  \renewcommand{\arraystretch}{1.15}
  \resizebox{0.9\linewidth}{!}{
  \begin{tabular}{l|l|cccc}
    \toprule
    Model &Type &ROUGE-L & BLEU-1 & BLEU-4 & gFVD$\downarrow$  \\
    \midrule
    Llama-3.2~\cite{grattafiori2024llama}&  AR    &  20.25 & 19.96  & 3.65  & 1579.86    \\
    LLaDA~\cite{nie2025large}&   MDM& 24.73  & 23.14  & 6.20 &  1306.30  \\
     \baseline{VQ-Diffusion} & \baseline{\textbf{VQD}}&     
     \baseline{\textbf{27.29}} 
    & \baseline{\textbf{26.55}} 
    & \baseline{\textbf{7.91}} 
    & \baseline{\textbf{1107.46}}  \\
    \bottomrule
  \end{tabular} } 
    \end{minipage}\hfill
    \begin{minipage}[t]{0.45\linewidth}
    \centering 
  \setlength{\tabcolsep}{4pt}
  \renewcommand{\arraystretch}{1.15}
     \captionof{table}{  \label{tab:gen_paradigms} Comparison of representation paradigms.}
  \resizebox{0.9\linewidth}{!}{
  \begin{tabular}{c|ccccc}
    \toprule
    Model & ROUGE-L & BLEU-1 & BLEU-4 & gFVD$\downarrow$  \\
    \midrule
    Continuous    & 26.10  & 25.74  & 7.65 &  1194.32  \\
     \baseline{Discrete} &   
     \baseline{\textbf{27.29}} 
    & \baseline{\textbf{26.55}} 
    & \baseline{\textbf{7.91}} 
    & \baseline{\textbf{1107.46}} \\
    \bottomrule
  \end{tabular} } 
    \end{minipage}  
\end{figure*}

%% file: table/model_para.tex
\begin{wraptable}{r}{0.49\linewidth}
    \centering
    \vspace{-10pt}
    \caption{Comparison of training strategies for the LLM and diffusion components.}
    \label{tab:training}
    \vspace{-5pt}
    \setlength{\tabcolsep}{4pt}
    \renewcommand{\arraystretch}{1.15}
    \resizebox{\linewidth}{!}{
    \begin{tabular}{l|cccc}
        \toprule
        Training strategy & ROUGE-L & BLEU-1 & BLEU-4 & gFVD$\downarrow$ \\
        \midrule
        Separate & 24.09 & 23.86 & 6.36 & 1373.61 \\
        \baseline{Joint} 
        & \baseline{\textbf{27.29}}
        & \baseline{\textbf{26.55}}
        & \baseline{\textbf{7.91}}
        & \baseline{\textbf{1107.46}} \\
        \bottomrule
    \end{tabular}}
    \vspace{-10pt}
\end{wraptable}
 

%% file: sec/6_conclusion.tex
\section{Conclusion}
In this paper, we present a first end-to-end SLVG framework: LLMCond, which  generates SL videos directly from text without relying on intermediate modalities. Specifically, to align textual input with the sign sequence, we introduce an LLM-based conditioner trained with a CTC alignment, to produce semantically consistent clip-level conditions. To ensure temporal coherence across clips, we employ a GCR that smooths the generated conditions over time. To enable efficient yet expressive video synthesis, we compress videos into discrete tokens with the SignVQ, and generate latent video tokens using a VQ-diffusion model conditioned on the refined conditions. Extensive experiments on public SL datasets demonstrate that LLMCond delivers highly competitive performance, producing temporally coherent and visually expressive videos without relying on extra visual modalities such as pose, depth, or optical flow.

%% file: sec/sup.tex
\setcounter{table}{8}
\setcounter{figure}{6}

\input{table/BTpose}

\section{More Comparison}

\paragraph{Comparison with text-to-pose SLP models.}
Although our goal is text-to-video generation, we evaluate the accuracy of the generated signs by extracting pose sequences from the generated videos using HRNet and measuring their correctness. 
We follow ProTran~\cite{saunders2020progressive} and train a SLT model~\cite{camgoz2020sign} for pose back-translation.

As shown in Table~\ref{tab:pose_back_translation_compare}, we compare our method with existing text-to-pose sign language production models. LLMCond consistently outperforms previous SLP methods (\eg ProTran ~\cite{saunders2020progressive}) across all evaluation metrics. Notably, the SLP baselines generate pose sequences directly from text, making their outputs naturally compatible with our pose back-translation evaluation. In comparison, LLMCond must first produce videos from which poses are then extracted, introducing potential errors that could influence the evaluation. Nevertheless, these results further demonstrate LLMCond's ability to maintain semantic fidelity.

\begin{wrapfigure}{l}{0.5\textwidth} 
\includegraphics[width=\linewidth]{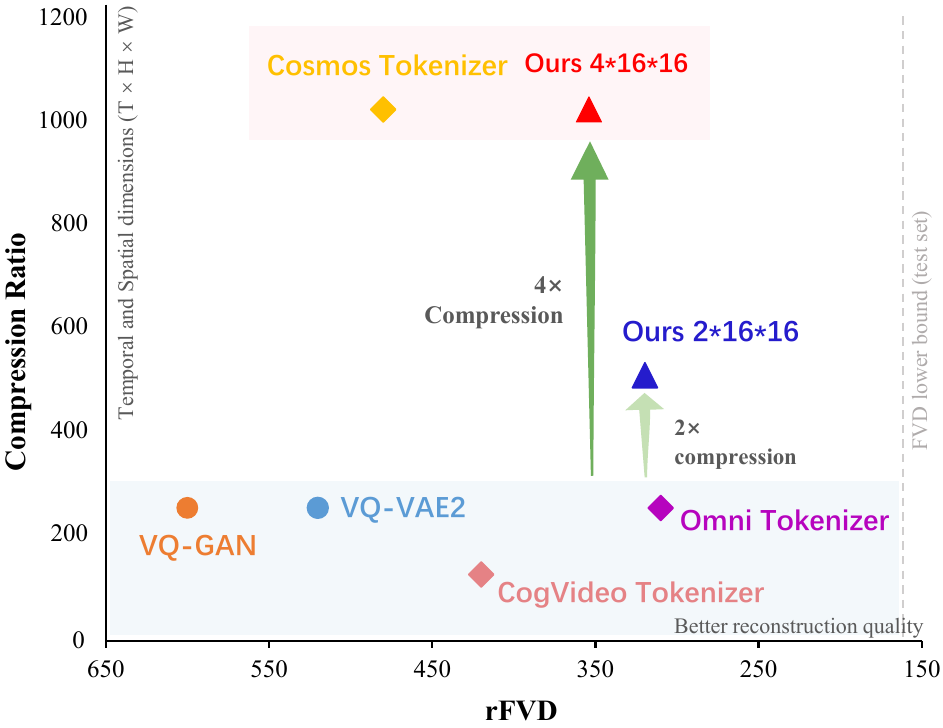}
\caption{Compression Ratio and rFVD Comparison across Different Tokenizers on Phoenix14T.}
\label{fig:comparevq} 
\end{wrapfigure}

\paragraph{Comparison of Video Tokenizer.}
We further compare our video tokenizer, SignVQ, with representative image/video tokenizers, including VQ-GAN~\cite{esser2021taming}, VQ-VAE2~\cite{razavi2019generating}, Cosmos Tokenizer~\cite{Agarwal2025Cosmos}, Omni Tokenizer~\cite{Wang2024OmniTokenizer}, and CogVideo Tokenizer~\cite{hong2022cogvideo}. As shown in Figure~\ref{fig:comparevq}, we present the trade-off between compression ratio and reconstruction quality measured by rFVD. Compared with VQ-GAN and VQ-VAE2, SignVQ achieves higher compression while maintaining strong reconstruction and generation performance. Relative to Omni Tokenizer and CogVideo Tokenizer, it achieves up to 1024× compression within an acceptable rFVD range. Compared with Cosmos Tokenizer, our method attains higher-quality reconstruction at similar compression levels.
Overall, these results demonstrate that our SignVQ tokenizer achieves a favorable balance between compactness and fidelity.  \\



\section{User Study}
In addition to objective metrics (\ie back-translation accuracy and FVD), we conducted a user study to evaluate the semantic comprehensibility and motion naturalness of the generated sign sequences. 
We randomly sampled 10 non-overlapping videos from each dataset (Phoenix14T and CSL-Daily), obtaining 20 unique videos in total. Each video was 2--8 seconds long, resized to 256×256 at 25 fps, and shown alongside its ground-truth reference. Ten participants took part in a within-subject design. For every dataset, each participant rated randomized, method-blind videos from our method and a baseline, presented in random order per video. For each video, raters scored two criteria on an overall 
\begin{wrapfigure}{r}{0.5\textwidth} 
\vspace{-5mm}
\includegraphics[width=\linewidth]{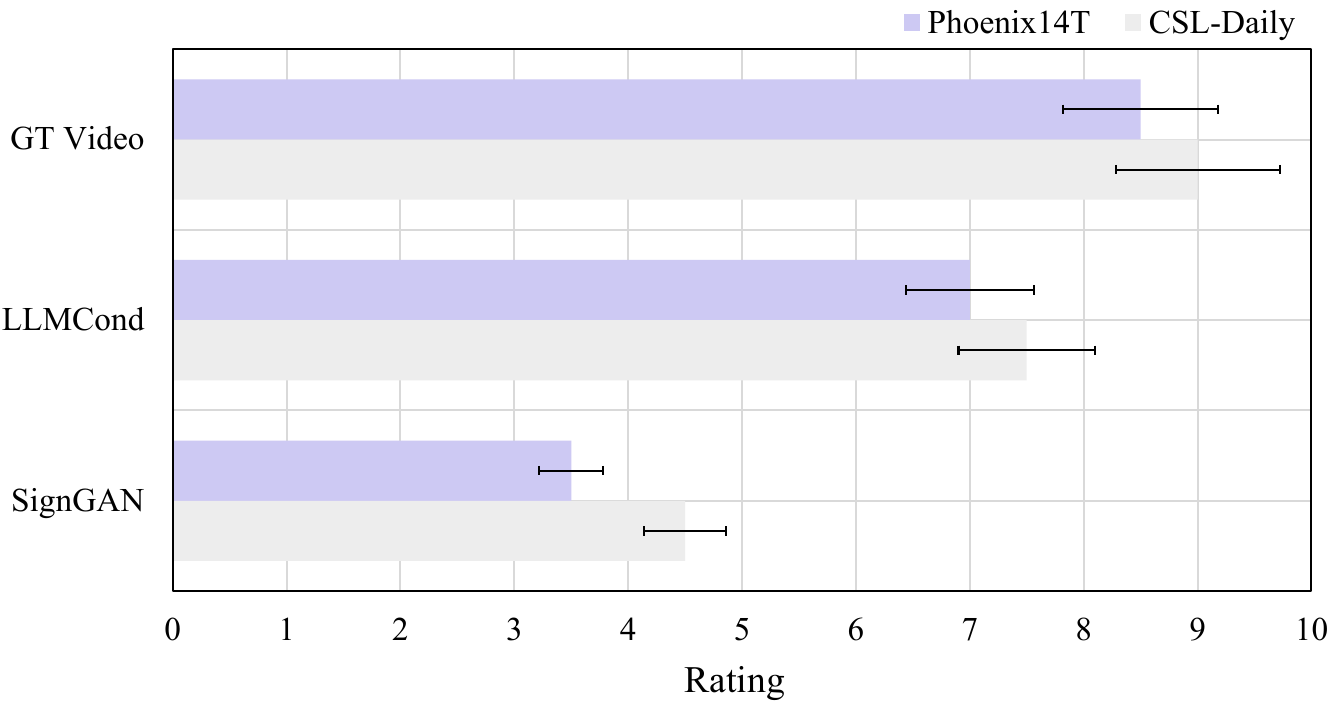}
\caption{Average user ratings across semantic consistency and motion naturalness on Phoenix14T and CSL-Daily.}
\vspace{-3mm}
\label{fig:user} 
\end{wrapfigure}
10-point rating scale: 
(i) semantic consistency with the associated text and (ii) motion naturalness. Video order and method labels were randomized for every participant, and no additional instructions were provided.

As shown in Figure~\ref{fig:user}, ground-truth videos receive the highest ratings on both datasets, forming the upper bound. Our method (LLMCond) obtains higher scores than SignGAN, with an average improvement of roughly +3.5 points on Phoenix14T and +3 points on CSL-Daily. These results confirm that LLMCond produces more semantically faithful and motion-accurate sign sequences than the baseline.

\section{Architecture Details}\label{sec:details}

\begin{wrapfigure}{r}{0.6\textwidth} 
\centering 
    \includegraphics[width=\linewidth]{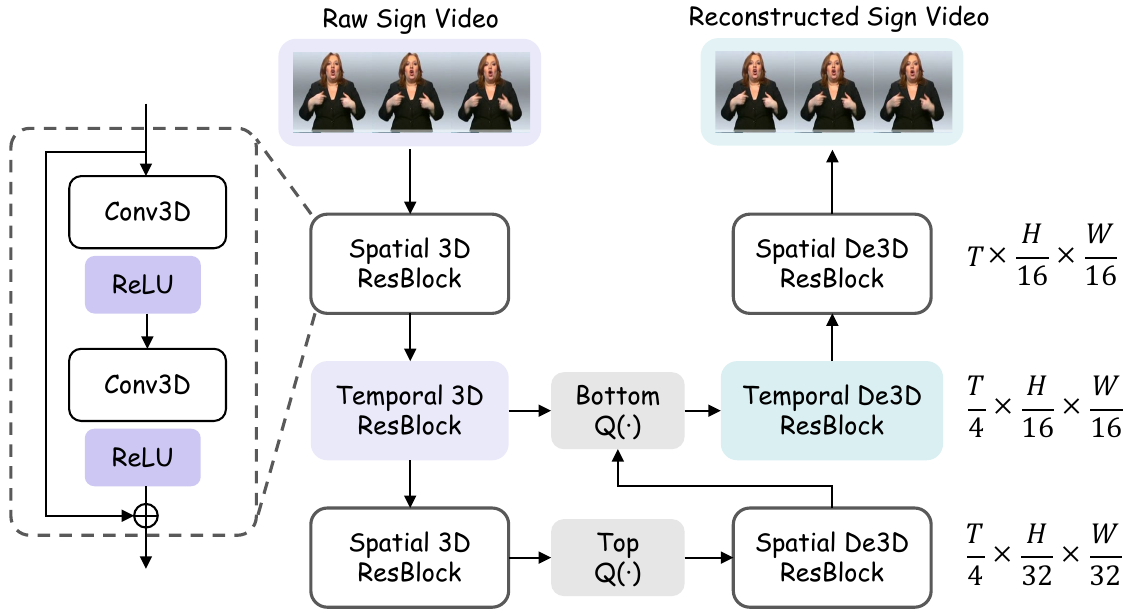}
    \vspace{-3mm}
   \caption{The architecture of the proposed SignVQ tokenizer. Left: encoder that downsamples the sign video into top and bottom discrete latents. Right: decoder that upsamples these latents with De3D ResBlocks to reconstruct the video.}
   \label{fig:vqvae} 
\vspace{-3mm}
\end{wrapfigure}

\textbf{SignVQ.} 
As shown in Figure~\ref{fig:vqvae}, the encoder of SignVQ is composed of seven 3D ResNet blocks arranged in a spatial--temporal--spatial order. 
The first four blocks are spatial 3D ResNet blocks (kernel size $1\times3\times3$) that progressively downsample the feature maps along the height and width dimensions, resulting in an overall $16\times$ reduction in spatial resolution. 
They are followed by two temporal ResNet blocks ($3\times1\times1$), which further downsample the video by a factor of $4$ along the temporal dimension. 
Finally, an additional spatial 3D ResNet block ($1\times3\times3$) is employed to further reduce the spatial resolution and produce a compact latent representation that preserves high-level spatial semantics.
The decoder mirrors this spatial--temporal--spatial structure using De3D (3D deconvolutional) ResNet blocks. 
Starting from the quantized latents, a spatial De3D block operates at the coarsest resolution, two temporal De3D blocks upsample the sequence in time, and the final four spatial De3D blocks upsample the feature maps in space to reconstruct the full-resolution video. 

\textbf{LLM Conditioner.} The LLM Conditioner is initialized with the pretrained weights of the Llama 3.2 1B model from HuggingFace, to fully leverage the rich linguistic knowledge of large language models. We apply LoRA adaptation with a rank of 64 and a scaling factor ($\alpha$) of 32.

\textbf{VQ-Diffusion.} Following the open-sourced VQ-Diffusion, we adopt the pretrained VQ-Diffusion weights on ImageNet~\footnote{\url{https://github.com/microsoft/VQ-Diffusion/tree/main}} for initialization.

\textbf{Reference Image Encoder.}
We adopt a ResNet-18 backbone as the reference image encoder to extract reference image embeddings. During training, the reference frame is randomly selected from the input frame sequence, while during inference, it is fixed as the first frame of the generated sign language video.

\textbf{Condition2Visual Alignment.}
Before the condition embeddings are fed into the VQ-Diffusion model, a Condition2Visual Mapper is employed to align feature dimensions between the LLM Conditioner and the diffusion model. The Condition2Visual Mapper consists of two MLP layers with a ReLU activation in between. We omit its description from the main paper for clarity of presentation.

\textbf{Video Length Prediction.}
We train a lightweight model to predict the target video length based on the text embeddings. The Video Length Predictor takes the output embeddings from the Llama embedding layer as input and consists of four Transformer encoder layers followed by an MLP layer to estimate the final video length. We omit its description from the main paper for clarity of presentation.

\paragraph{Training Details.}
The entire implementation is based on PyTorch 2.1.\\
\textit{(1) SignVQ.} The SignVQ module is trained on a single NVIDIA A6000 GPU with 48 GB for 50 epochs using the Adam optimizer with an initial learning rate of $1\times10^{-4}$. A MultiStepLR scheduler is applied with a decay factor of 0.5 at epochs 1, 3, 10, 30, and 35.\\
\textit{(2) LLM Conditioner and VQ-Diffusion.}
The LLM Conditioner and VQ-Diffusion modules are trained jointly using Adam~\cite{kingma2014adam} with an initial learning rate of $1\times10^{-5}$, a batch size of $3$, and $20$k training steps per stage. The learning rate follows a MultiStepLR schedule with a decay factor of 0.5 at epochs 3, 10, and 30. All experiments are conducted on four NVIDIA A6000 GPUs. For generation, we follow the standard VQ-Diffusion framework~\cite{gu2022vector} with classifier-free guidance~\cite{ho2022classifier} (guidance scale set to $3.5$) and use $100$ denoising steps during inference. \\
\textit{(3) Back-Translation Model.}
To evaluate the semantic consistency of the generated sign language videos, we adopt the SLT model~\cite{gan2025mixsigngraph} to translate sign language videos (\ie the video back-translation model) into texts. 
The video back-translation model is trained using a single NVIDIA RTX A6000 GPU. An Adam optimizer~\cite{kingma2014adam} is utilized with a learning rate of 1e-3, which decays by a factor of 0.5 every 10 epochs, and the batch size is set to 8.

\section{More Qualitative Results}
\begin{figure*}[t]
  \centering
    \includegraphics[width=1.0\linewidth]{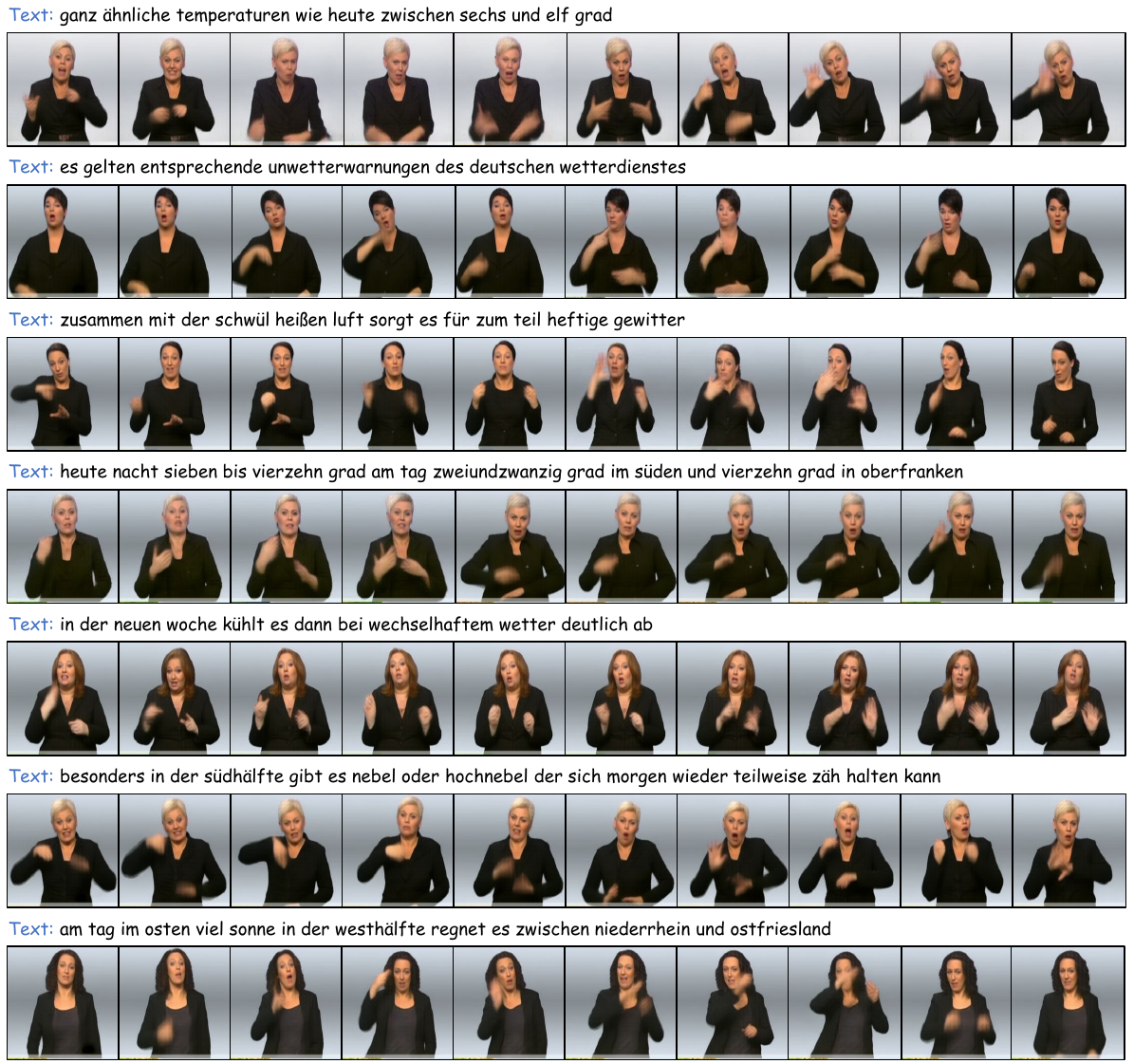}
    \vspace{-4mm}
   \caption{More generated cases of Phoenix14T dataset.}
   \label{fig:sup_vis1} 
  \vspace{-1mm}
\end{figure*}

\begin{figure*}[t]
  \centering
    \includegraphics[width=1.0\linewidth]{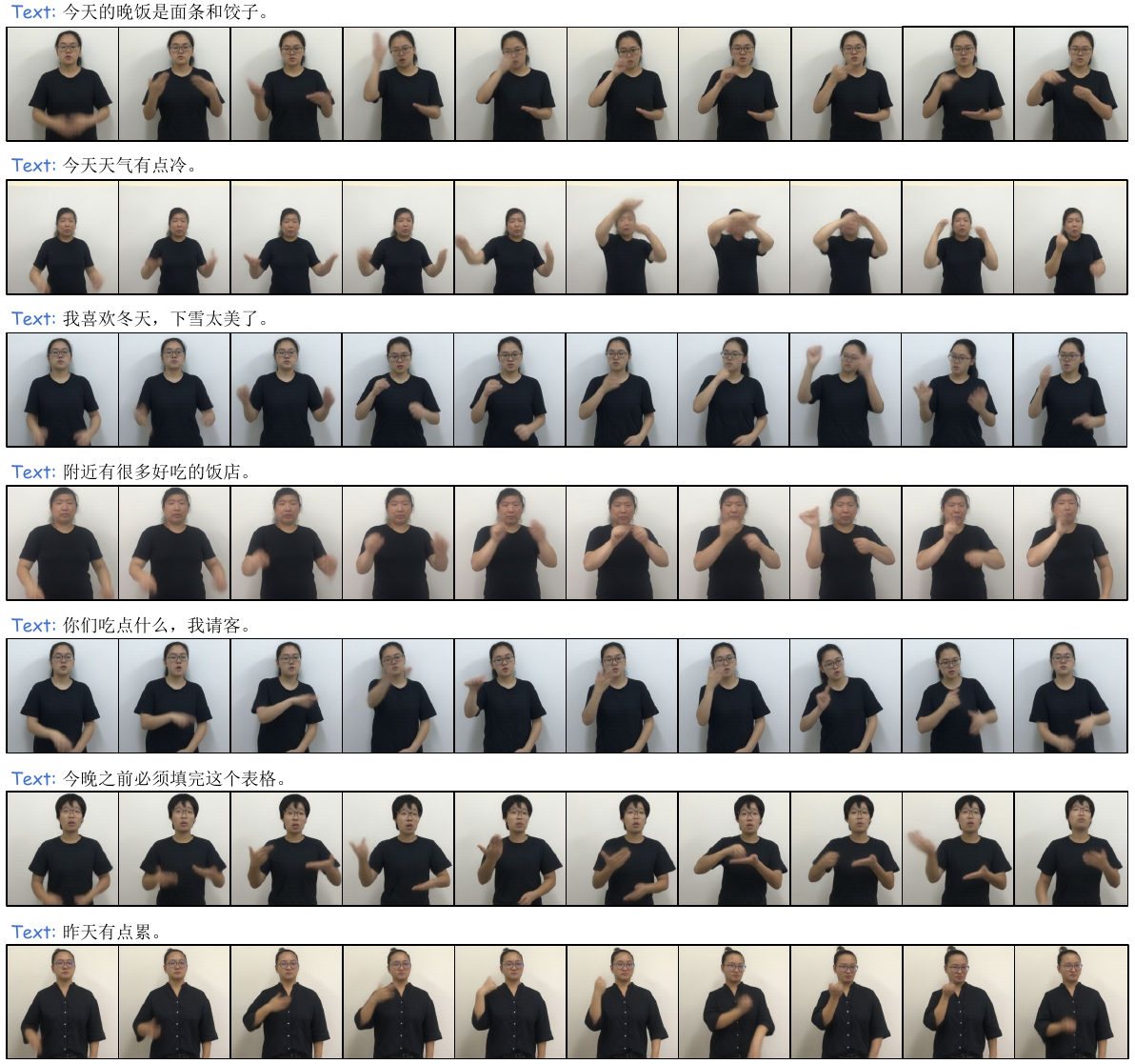}
    \vspace{-4mm}
   \caption{More generated cases of CSL-Daily dataset.}
   \label{fig:sup_vis2} 
  \vspace{-1mm}
\end{figure*}

We present additional qualitative examples to further demonstrate the effectiveness of our proposed method.
As illustrated in Figure~\ref{fig:sup_vis1} and Figure~\ref{fig:sup_vis2}, we provide more generated samples from both the Phoenix14T and CSL-Daily datasets. 


\section{Limitations and Broader impacts}
\label{sec:limitation}
In this paper, we propose a novel end-to-end SLVG model, LLMCond, which achieves notable performance improvements. Nevertheless, our framework still has two limitations that warrant further investigation.
The first limitation lies in its data-driven nature. Since the model learns patterns directly from the training data, it may inherit dataset biases and occasionally produce uncontrolled or inconsistent video content. Ensuring careful dataset curation and maintaining a balanced data distribution are therefore crucial to mitigating such issues.
The second limitation lies in generalization. Although our model performs well on benchmark datasets, its robustness to varying lighting conditions and diverse signing styles remains to be further validated.

This paper presents work whose goal is to advance the field of machine learning for sign language video generation. By improving end-to-end generation of sign language videos, our work has the potential to support the development of more accessible communication technologies for the Hard-of-Hearing community. At the same time, we acknowledge that sign language technologies may inherit biases present in training data, including variations across signers, regions, and signing styles. While our work focuses on modeling and representation learning rather than deployment, responsible use of such systems should consider data diversity, fairness, and appropriate human oversight. Overall, we believe the primary impact of this work is positive, contributing to more inclusive and efficient sign language research, and we do not foresee significant negative societal consequences beyond those commonly associated with machine learning research.

%% file: table/BTpose.tex
\begin{table}[t]
\centering
\caption{Comparison of \textbf{pose back-translation} performance. We additionally report DTW-PA-MPJPE, computed by applying DTW to the per-frame Procrustes-aligned joint position error sequence to obtain a sequence-level distance robust to temporal warping.}
\tablestyle{1pt}{1.1}
\resizebox{1.0\linewidth}{!}{
\begin{tabular}{l|cccccc|cccccc}
\toprule
\multirow{2}{*}{Method}
& \multicolumn{6}{c|}{Phoenix14T}
& \multicolumn{6}{c}{CSL-Daily} \\
\cline{2-13}
& BLEU1 & BLEU2 & BLEU3 & BLEU4 & ROUGE & DTW$\downarrow$
& BLEU1 & BLEU2 & BLEU3 & BLEU4 & ROUGE & DTW$\downarrow$ \\
\midrule
Ground-Truth
& 30.99 & 18.36 & 12.83 & 9.87 & 31.02 & 0
& 31.54 & 19.32 & 13.54 & 9.91 & 32.73 & 0 \\
\midrule
ProTran~\cite{saunders2020progressive}
& 17.96 & 8.99 & 5.64 & 4.07 & 20.97 & 16.80
& 14.85 & 8.25 & 6.31 & 4.85 & 17.02 & 19.57 \\
Adversarial~\cite{saunders2020adversarial}
& 17.70 & 8.96 & 5.72 & 4.18 & 21.15 & 16.43
& 14.69 & 8.13 & 6.55 & 4.26 & 16.94 & 18.81 \\
MDN~\cite{saunders2021continuous}
& 18.06 & 9.30 & 6.06 & 4.52 & 21.44 & 14.16
& 16.17 & 10.52 & 7.87 & 5.66 & 18.41 & 15.20 \\
MoMP~\cite{saunders2021mixed}
& 20.55 & 10.98 & 7.02 & 5.14 & 20.75 & 12.31
& 19.27 & 11.37 & 7.95 & 5.94 & 22.16 & 12.53 \\
SignGen~\cite{qi2024signgen}
& 10.42 & 2.42 & 0.89 & 0.38 & 12.68 & 44.60
& 9.36 & 1.86 & 0.41 & 0.02 & 11.33 & 43.95 \\
\baseline{LLMCond}
& \baseline{\textbf{22.70}} & \baseline{\textbf{11.26}} & \baseline{\textbf{7.04}} & \baseline{\textbf{5.59}} & \baseline{\textbf{23.39}} & \baseline{\textbf{9.41}}
& \baseline{\textbf{22.56}} & \textbf{\baseline{13.96}} & \baseline{\textbf{9.35}} & \baseline{\textbf{6.52}} & \baseline{\textbf{23.83}} & \baseline{\textbf{8.78}} \\
\bottomrule
\end{tabular}}
\vspace{-4mm}
\label{tab:pose_back_translation_compare}
\end{table}

%% file: checklist.tex
\section*{NeurIPS Paper Checklist}

\begin{enumerate}

\item {\bf Claims}
    \item[] Question: Do the main claims made in the abstract and introduction accurately reflect the paper's contributions and scope?
        \item[] Answer: \answerYes{} 
    \item[] Justification: {The abstract and introduction clearly summarize the main contributions of the paper, including the proposal of LLM-based conditioner and the end-to-end SLVG models. (see Section~\ref{sec:intrduction}).}

\item {\bf Limitations}
    \item[] Question: Does the paper discuss the limitations of the work performed by the authors?
    \item[] Answer: \answerYes{} 
    \item[] Justification:  We include a dedicated ``Limitations and Broader impacts '' section in the appendix of main paper, where we discuss practical challenges, and the potential negative social impacts including high resource demands, data bias, and model reliability in real-world applications. (see Section~\ref{sec:limitation}).

\item {\bf Theory assumptions and proofs}
    \item[] Question: For each theoretical result, does the paper provide the full set of assumptions and a complete (and correct) proof?
    \item[] Answer: \answerNA{} 
    \item[] Justification: Our work focuses on empirical model design and evaluation for sign language video generation. It does not include theoretical results that require formal assumptions or proofs.

    \item {\bf Experimental result reproducibility}
    \item[] Question: Does the paper fully disclose all the information needed to reproduce the main experimental results of the paper to the extent that it affects the main claims and/or conclusions of the paper (regardless of whether the code and data are provided or not)?
      \item[] Answer: \answerYes{} 
    \item[] Justification:  We provide comprehensive details of the model architecture, training settings, and evaluation protocol in Section~\ref{sec:experiment}. Specifically, we describe the datasets used, data preprocessing steps, model configurations, hyperparameters, and optimization strategies. We also report implementation details such as training schedules, batch sizes, and evaluation metrics to ensure reproducibility. Furthermore, additional model details and implementation specifications are provided in the supplementary materials. We will also release the code to further facilitate reproducibility.

\item {\bf Open access to data and code}
    \item[] Question: Does the paper provide open access to the data and code, with sufficient instructions to faithfully reproduce the main experimental results, as described in supplemental material?
  \item[] Answer: \answerNo{} 
    \item[] Justification: The paper and supplementary material provide sufficient implementation details, training settings, and experimental configurations to facilitate reproduction of the main results.  

\item {\bf Experimental setting/details}
    \item[] Question: Does the paper specify all the training and test details (e.g., data splits, hyperparameters, how they were chosen, type of optimizer) necessary to understand the results?
    \item[] Answer: \answerYes{} 
    \item[] Justification: The paper provides comprehensive training and testing details, including data splits, model architecture, hyperparameters, learning rate, optimizer, and training epochs.  

\item {\bf Experiment statistical significance}
    \item[] Question: Does the paper report error bars suitably and correctly defined or other appropriate information about the statistical significance of the experiments?
     \item[] Answer: \answerNo{}{} 
    \item[] Justification: While we report the main experimental results, we do not include error bars or statistical significance tests due to limited computational resources. In future work, we plan to provide more extensive statistical analysis over multiple random seeds or cross-validation splits.

\item {\bf Experiments compute resources}
    \item[] Question: For each experiment, does the paper provide sufficient information on the computer resources (type of compute workers, memory, time of execution) needed to reproduce the experiments?
   \item[] Answer: \answerYes{} 
    \item[] Justification: The paper specifies the type of computing resources used, including GPU model (e.g.,  A6000 GPU), memory (e.g., 48GB) (See Section~\ref{sec:details}). 
    
\item {\bf Code of ethics}
    \item[] Question: Does the research conducted in the paper conform, in every respect, with the NeurIPS Code of Ethics \url{https://neurips.cc/public/EthicsGuidelines}?
     \item[] Answer:  \answerYes{} 
    \item[] Justification: The research conforms in every respect with the NeurIPS Code of Ethics. We have ensured transparency, reproducibility, and fairness in our methodology and reporting.

\item {\bf Broader impacts}
    \item[] Question: Does the paper discuss both potential positive societal impacts and negative societal impacts of the work performed?
   \item[] Answer: \answerYes{}{} 
    \item[] Justification:  Our work discusses both potential positive and negative societal impacts.(See Section~\ref{sec:limitation})
    
\item {\bf Safeguards}
    \item[] Question: Does the paper describe safeguards that have been put in place for responsible release of data or models that have a high risk for misuse (e.g., pre-trained language models, image generators, or scraped datasets)?
    \item[] Answer: \answerNA{} 
    \item[] Justification: The model weights used in this work are trained solely on publicly available datasets, and we do not release any new datasets or pretrained models that pose a high risk for misuse. Therefore, no additional safeguards are necessary.

\item {\bf Licenses for existing assets}
    \item[] Question: Are the creators or original owners of assets (e.g., code, data, models), used in the paper, properly credited and are the license and terms of use explicitly mentioned and properly respected?
   \item[] Answer: \answerYes{}
    \item[] Justification: We used publicly available datasets and pretrained model weights, all of which are properly cited in the paper. The licenses and terms of use  for these assets were carefully reviewed and respected. We have also specified the sources, versions, and relevant URLs where appropriate.

\item {\bf New assets}
    \item[] Question: Are new assets introduced in the paper well documented and is the documentation provided alongside the assets?
   \item[] Answer: \answerNA{} 
    \item[] Justification: This paper does not release any new datasets, models, or code assets. All experiments are based on publicly available datasets and pretrained models, which are properly cited.

\item {\bf Crowdsourcing and research with human subjects}
    \item[] Question: For crowdsourcing experiments and research with human subjects, does the paper include the full text of instructions given to participants and screenshots, if applicable, as well as details about compensation (if any)? 
   \item[] Answer:  \answerYes{} 
    \item[] Justification: Appendix B provides the user study protocol, participant instructions, evaluation criteria, and compensation details.

\item {\bf Institutional review board (IRB) approvals or equivalent for research with human subjects}
    \item[] Question: Does the paper describe potential risks incurred by study participants, whether such risks were disclosed to the subjects, and whether Institutional Review Board (IRB) approvals (or an equivalent approval/review based on the requirements of your country or institution) were obtained?
    \item[] Answer: \answerYes{}
    \item[] Justification:  The user study underwent an equivalent internal ethics review in accordance with applicable institutional requirements.

\item {\bf Declaration of LLM usage}
    \item[] Question: Does the paper describe the usage of LLMs if it is an important, original, or non-standard component of the core methods in this research? Note that if the LLM is used only for writing, editing, or formatting purposes and does \emph{not} impact the core methodology, scientific rigor, or originality of the research, declaration is not required.
    \item[] Answer: \answerYes{}
    \item[] Justification: LLMs are a core component of our method, serving as a condition generator that transforms text into structured signals for sign language video generation. Their role is explicitly described in both the main paper and supplementary materials.

\end{enumerate}

%% file: neurips_2026.bbl
\begin{thebibliography}{56}
\providecommand{\natexlab}[1]{#1}
\providecommand{\url}[1]{\texttt{#1}}
\expandafter\ifx\csname urlstyle\endcsname\relax
  \providecommand{\doi}[1]{doi: #1}\else
  \providecommand{\doi}{doi: \begingroup \urlstyle{rm}\Url}\fi

\bibitem[Agarwal et~al.(2025)Agarwal, Ali, Bala, Balaji, Barker, Cai,
  Chattopadhyay, Chen, Cui, Ding, et~al.]{Agarwal2025Cosmos}
Niket Agarwal, Arslan Ali, Maciej Bala, Yogesh Balaji, Erik Barker, Tiffany
  Cai, Prithvijit Chattopadhyay, Yongxin Chen, Yin Cui, Yifan Ding, et~al.
\newblock Cosmos world foundation model platform for physical ai.
\newblock \emph{arXiv preprint arXiv:2501.03575}, 2025.

\bibitem[Baltatzis et~al.(2024)Baltatzis, Potamias, Ververas, Sun, Deng, and
  Zafeiriou]{baltatzis2024neural}
Vasileios Baltatzis, Rolandos~Alexandros Potamias, Evangelos Ververas,
  Guanxiong Sun, Jiankang Deng, and Stefanos Zafeiriou.
\newblock Neural sign actors: A diffusion model for 3d sign language production
  from text.
\newblock In \emph{Proceedings of the IEEE/CVF Conference on Computer Vision
  and Pattern Recognition}, pages 1985--1995, 2024.

\bibitem[Camgoz et~al.(2018)Camgoz, Hadfield, Koller, Ney, and
  Bowden]{camgoz2018neural}
Necati~Cihan Camgoz, Simon Hadfield, Oscar Koller, Hermann Ney, and Richard
  Bowden.
\newblock Neural sign language translation.
\newblock In \emph{Proceedings of the IEEE conference on computer vision and
  pattern recognition}, pages 7784--7793, 2018.

\bibitem[Camgoz et~al.(2020)Camgoz, Koller, Hadfield, and
  Bowden]{camgoz2020sign}
Necati~Cihan Camgoz, Oscar Koller, Simon Hadfield, and Richard Bowden.
\newblock Sign language transformers: Joint end-to-end sign language
  recognition and translation.
\newblock In \emph{Proceedings of the IEEE/CVF conference on computer vision
  and pattern recognition}, pages 10023--10033, 2020.

\bibitem[Chen et~al.(2023)Chen, Xia, He, Zhang, Cun, Yang, Xing, Liu, Chen,
  Wang, et~al.]{chen2023videocrafter1}
Haoxin Chen, Menghan Xia, Yingqing He, Yong Zhang, Xiaodong Cun, Shaoshu Yang,
  Jinbo Xing, Yaofang Liu, Qifeng Chen, Xintao Wang, et~al.
\newblock Videocrafter1: Open diffusion models for high-quality video
  generation.
\newblock \emph{arXiv preprint arXiv:2310.19512}, 2023.

\bibitem[Chen et~al.(2024)Chen, Zhang, Cun, Xia, Wang, Weng, and
  Shan]{chen2024videocrafter2}
Haoxin Chen, Yong Zhang, Xiaodong Cun, Menghan Xia, Xintao Wang, Chao Weng, and
  Ying Shan.
\newblock Videocrafter2: Overcoming data limitations for high-quality video
  diffusion models.
\newblock In \emph{Proceedings of the IEEE/CVF Conference on Computer Vision
  and Pattern Recognition}, pages 7310--7320, 2024.

\bibitem[Chen et~al.(2022)Chen, Zuo, Wei, Wu, Liu, and Mak]{chen2022two}
Yutong Chen, Ronglai Zuo, Fangyun Wei, Yu~Wu, Shujie Liu, and Brian Mak.
\newblock Two-stream network for sign language recognition and translation.
\newblock \emph{Advances in Neural Information Processing Systems},
  35:\penalty0 17043--17056, 2022.

\bibitem[Esser et~al.(2021)Esser, Rombach, and Ommer]{esser2021taming}
Patrick Esser, Robin Rombach, and Bjorn Ommer.
\newblock Taming transformers for high-resolution image synthesis.
\newblock In \emph{Proceedings of the IEEE/CVF conference on computer vision
  and pattern recognition}, pages 12873--12883, 2021.

\bibitem[Fang et~al.(2023)Fang, Sui, Zhang, and Tian]{fang2023signdiff}
Sen Fang, Chunyu Sui, Xuedong Zhang, and Yapeng Tian.
\newblock Signdiff: Learning diffusion models for american sign language
  production.
\newblock \emph{arXiv e-prints}, pages arXiv--2308, 2023.

\bibitem[Fang et~al.(2025)Fang, Chen, Wang, Zheng, Sui, and
  Tian]{fang2025signllm}
Sen Fang, Chen Chen, Lei Wang, Ce~Zheng, Chunyu Sui, and Yapeng Tian.
\newblock Signllm: Sign language production large language models.
\newblock In \emph{2025 IEEE/CVF International Conference on Computer Vision
  Workshops (ICCVW)}, pages 6681--6693. IEEE, 2025.

\bibitem[Gan et~al.(2024)Gan, Yin, Jiang, Wen, Xie, and Lu]{gan2024signgraph}
Shiwei Gan, Yafeng Yin, Zhiwei Jiang, Hongkai Wen, Lei Xie, and Sanglu Lu.
\newblock Signgraph: A sign sequence is worth graphs of nodes.
\newblock In \emph{Proceedings of the IEEE/CVF Conference on Computer Vision
  and Pattern Recognition}, pages 13470--13479, 2024.

\bibitem[Gan et~al.(2025)Gan, Yin, Jiang, Xie, Lu, and
  Wen]{gan2025mixsigngraph}
Shiwei Gan, Yafeng Yin, Zhiwei Jiang, Lei Xie, Sanglu Lu, and Hongkai Wen.
\newblock Mixsigngraph: A sign sequence is worth mixed graphs of nodes.
\newblock In \emph{The Thirty-ninth Annual Conference on Neural Information
  Processing Systems}, 2025.

\bibitem[Gan et~al.(2026{\natexlab{a}})Gan, Liu, Yin, Jiang, Guo, Xie, Lu, and
  Wen]{gan2026signllama}
Shiwei Gan, Xiao Liu, Yafeng Yin, Zhiwei Jiang, Bowen Guo, Lei Xie, Sanglu Lu,
  and Hongkai Wen.
\newblock Signllama: Enhancing gloss-free sign language translation by
  prioritizing visual features for llms.
\newblock \emph{arXiv preprint arXiv:2608.09006}, 2026{\natexlab{a}}.

\bibitem[Gan et~al.(2026{\natexlab{b}})Gan, Wang, Liu, Yin, Liu, Lu, and
  Xie]{gan2026sign}
Shiwei Gan, Lichen Wang, Xiao Liu, Yafeng Yin, Kuizhuang Liu, Sanglu Lu, and
  Lei Xie.
\newblock Sign language question answering: A new task, benchmark, and baseline
  for sign language understanding.
\newblock \emph{arXiv preprint arXiv:2607.27826}, 2026{\natexlab{b}}.

\bibitem[Grattafiori et~al.(2024)Grattafiori, Dubey, Jauhri, Pandey, Kadian,
  Al-Dahle, Letman, Mathur, Schelten, Vaughan, et~al.]{grattafiori2024llama}
Aaron Grattafiori, Abhimanyu Dubey, Abhinav Jauhri, Abhinav Pandey, Abhishek
  Kadian, Ahmad Al-Dahle, Aiesha Letman, Akhil Mathur, Alan Schelten, Alex
  Vaughan, et~al.
\newblock The llama 3 herd of models.
\newblock \emph{arXiv preprint arXiv:2407.21783}, 2024.

\bibitem[Gu et~al.(2022)Gu, Chen, Bao, Wen, Zhang, Chen, Yuan, and
  Guo]{gu2022vector}
Shuyang Gu, Dong Chen, Jianmin Bao, Fang Wen, Bo~Zhang, Dongdong Chen, Lu~Yuan,
  and Baining Guo.
\newblock Vector quantized diffusion model for text-to-image synthesis.
\newblock In \emph{Proceedings of the IEEE/CVF conference on computer vision
  and pattern recognition}, pages 10696--10706, 2022.

\bibitem[Guo et~al.(2023)Guo, Yang, Rao, Liang, Wang, Qiao, Agrawala, Lin, and
  Dai]{guo2023animatediff}
Yuwei Guo, Ceyuan Yang, Anyi Rao, Zhengyang Liang, Yaohui Wang, Yu~Qiao,
  Maneesh Agrawala, Dahua Lin, and Bo~Dai.
\newblock Animatediff: Animate your personalized text-to-image diffusion models
  without specific tuning.
\newblock \emph{arXiv preprint arXiv:2307.04725}, 2023.

\bibitem[Heusel et~al.(2017)Heusel, Ramsauer, Unterthiner, Nessler, and
  Hochreiter]{heusel2017gans}
Martin Heusel, Hubert Ramsauer, Thomas Unterthiner, Bernhard Nessler, and Sepp
  Hochreiter.
\newblock Gans trained by a two time-scale update rule converge to a local nash
  equilibrium.
\newblock \emph{Advances in neural information processing systems}, 30, 2017.

\bibitem[Ho and Salimans(2022)]{ho2022classifier}
Jonathan Ho and Tim Salimans.
\newblock Classifier-free diffusion guidance.
\newblock \emph{arXiv preprint arXiv:2207.12598}, 2022.

\bibitem[Hong et~al.(2022)Hong, Ding, Zheng, Liu, and Tang]{hong2022cogvideo}
Wenyi Hong, Ming Ding, Wendi Zheng, Xinghan Liu, and Jie Tang.
\newblock Cogvideo: Large-scale pretraining for text-to-video generation via
  transformers.
\newblock \emph{arXiv preprint arXiv:2205.15868}, 2022.

\bibitem[Hu et~al.(2022)Hu, Shen, Wallis, Allen-Zhu, Li, Wang, Wang, Chen,
  et~al.]{hu2022lora}
Edward~J Hu, Yelong Shen, Phillip Wallis, Zeyuan Allen-Zhu, Yuanzhi Li, Shean
  Wang, Lu~Wang, Weizhu Chen, et~al.
\newblock Lora: Low-rank adaptation of large language models.
\newblock \emph{ICLR}, 1\penalty0 (2):\penalty0 3, 2022.

\bibitem[Hu et~al.(2021)Hu, Zhou, and Li]{hu2021hand}
Hezhen Hu, Wengang Zhou, and Houqiang Li.
\newblock Hand-model-aware sign language recognition.
\newblock In \emph{Proceedings of the AAAI conference on artificial
  intelligence}, volume~35, pages 1558--1566, 2021.

\bibitem[Hu(2024)]{hu2024animate}
Li~Hu.
\newblock Animate anyone: Consistent and controllable image-to-video synthesis
  for character animation.
\newblock In \emph{Proceedings of the IEEE/CVF Conference on Computer Vision
  and Pattern Recognition}, pages 8153--8163, 2024.

\bibitem[Huang et~al.(2018)Huang, Zhou, Li, and Li]{huang2018attention}
Jie Huang, Wengang Zhou, Houqiang Li, and Weiping Li.
\newblock Attention-based 3d-cnns for large-vocabulary sign language
  recognition.
\newblock \emph{IEEE Transactions on Circuits and Systems for Video
  Technology}, 29\penalty0 (9):\penalty0 2822--2832, 2018.

\bibitem[Kingma(2014)]{kingma2014adam}
Diederik~P Kingma.
\newblock Adam: A method for stochastic optimization.
\newblock \emph{arXiv preprint arXiv:1412.6980}, 2014.

\bibitem[Lin and Och(2004)]{lin2004automatic}
Chin-Yew Lin and Franz~Josef Och.
\newblock Automatic evaluation of machine translation quality using longest
  common subsequence and skip-bigram statistics.
\newblock In \emph{Proceedings of the 42nd annual meeting of the association
  for computational linguistics (ACL-04)}, pages 605--612, 2004.

\bibitem[Liu et~al.(2026{\natexlab{a}})Liu, Gan, Yin, Guo, Jiang, Meng, Xie,
  and Lu]{liu2026signpr}
Xiao Liu, Shiwei Gan, Yafeng Yin, Bowen Guo, Zhiwei Jiang, Shunmei Meng, Lei
  Xie, and Sanglu Lu.
\newblock Signpr: A progressive vector-quantized diffusion framework for sign
  language production.
\newblock In \emph{Proceedings of the IEEE/CVF Conference on Computer Vision
  and Pattern Recognition}, pages 2198--2208, 2026{\natexlab{a}}.

\bibitem[Liu et~al.(2026{\natexlab{b}})Liu, Gan, Yin, Yin, Guo, Sun, Jiang, and
  Xie]{liu2026bridging}
Xiao Liu, Shiwei Gan, Yafeng Yin, Jiaxin Yin, Bowen Guo, Yaqi Sun, Zhiwei
  Jiang, and Lei Xie.
\newblock Bridging the gap between semantics and reconstruction: Unifying sign
  language translation and production.
\newblock \emph{arXiv preprint arXiv:2608.09045}, 2026{\natexlab{b}}.

\bibitem[Nie et~al.(2025)Nie, Zhu, You, Zhang, Ou, Hu, Zhou, Lin, Wen, and
  Li]{nie2025large}
Shen Nie, Fengqi Zhu, Zebin You, Xiaolu Zhang, Jingyang Ou, Jun Hu, Jun Zhou,
  Yankai Lin, Ji-Rong Wen, and Chongxuan Li.
\newblock Large language diffusion models.
\newblock \emph{arXiv preprint arXiv:2502.09992}, 2025.

\bibitem[Papineni et~al.(2002)Papineni, Roukos, Ward, and
  Zhu]{papineni2002bleu}
Kishore Papineni, Salim Roukos, Todd Ward, and Wei-Jing Zhu.
\newblock Bleu: a method for automatic evaluation of machine translation.
\newblock In \emph{Proceedings of the 40th annual meeting of the Association
  for Computational Linguistics}, pages 311--318, 2002.

\bibitem[Qi et~al.(2024)Qi, Duan, Zhang, and Xu]{qi2024signgen}
Fan Qi, Yu~Duan, Huaiwen Zhang, and Changsheng Xu.
\newblock Signgen: End-to-end sign language video generation with latent
  diffusion.
\newblock In \emph{European Conference on Computer Vision}, pages 252--270.
  Springer, 2024.

\bibitem[Rastgoo et~al.(2021)Rastgoo, Kiani, Escalera, and
  Sabokrou]{rastgoo2021sign}
Razieh Rastgoo, Kourosh Kiani, Sergio Escalera, and Mohammad Sabokrou.
\newblock Sign language production: A review.
\newblock In \emph{Proceedings of the IEEE/CVF conference on computer vision
  and pattern recognition}, pages 3451--3461, 2021.

\bibitem[Razavi et~al.(2019)Razavi, Van~den Oord, and
  Vinyals]{razavi2019generating}
Ali Razavi, Aaron Van~den Oord, and Oriol Vinyals.
\newblock Generating diverse high-fidelity images with vq-vae-2.
\newblock \emph{Advances in neural information processing systems}, 32, 2019.

\bibitem[Saunders et~al.(2020{\natexlab{a}})Saunders, Camgoz, and
  Bowden]{saunders2020adversarial}
Ben Saunders, Necati~Cihan Camgoz, and Richard Bowden.
\newblock Adversarial training for multi-channel sign language production.
\newblock \emph{arXiv preprint arXiv:2008.12405}, 2020{\natexlab{a}}.

\bibitem[Saunders et~al.(2020{\natexlab{b}})Saunders, Camgoz, and
  Bowden]{saunders2020everybody}
Ben Saunders, Necati~Cihan Camgoz, and Richard Bowden.
\newblock Everybody sign now: Translating spoken language to photo realistic
  sign language video.
\newblock \emph{arXiv preprint arXiv:2011.09846}, 2020{\natexlab{b}}.

\bibitem[Saunders et~al.(2020{\natexlab{c}})Saunders, Camgoz, and
  Bowden]{saunders2020progressive}
Ben Saunders, Necati~Cihan Camgoz, and Richard Bowden.
\newblock Progressive transformers for end-to-end sign language production.
\newblock In \emph{European Conference on Computer Vision}, pages 687--705.
  Springer, 2020{\natexlab{c}}.

\bibitem[Saunders et~al.(2021{\natexlab{a}})Saunders, Camgoz, and
  Bowden]{saunders2021continuous}
Ben Saunders, Necati~Cihan Camgoz, and Richard Bowden.
\newblock Continuous 3d multi-channel sign language production via progressive
  transformers and mixture density networks.
\newblock \emph{International journal of computer vision}, 129\penalty0
  (7):\penalty0 2113--2135, 2021{\natexlab{a}}.

\bibitem[Saunders et~al.(2021{\natexlab{b}})Saunders, Camgoz, and
  Bowden]{saunders2021mixed}
Ben Saunders, Necati~Cihan Camgoz, and Richard Bowden.
\newblock Mixed signals: Sign language production via a mixture of motion
  primitives.
\newblock In \emph{Proceedings of the IEEE/CVF International Conference on
  Computer Vision}, pages 1919--1929, 2021{\natexlab{b}}.

\bibitem[Saunders et~al.(2022)Saunders, Camgoz, and
  Bowden]{saunders2022signing}
Ben Saunders, Necati~Cihan Camgoz, and Richard Bowden.
\newblock Signing at scale: Learning to co-articulate signs for large-scale
  photo-realistic sign language production.
\newblock In \emph{Proceedings of the IEEE/CVF Conference on Computer Vision
  and Pattern Recognition}, pages 5141--5151, 2022.

\bibitem[Shi et~al.(2022)Shi, Brentari, Shakhnarovich, and
  Livescu]{shi2022open}
Bowen Shi, Diane Brentari, Gregory Shakhnarovich, and Karen Livescu.
\newblock Open-domain sign language translation learned from online video.
\newblock In \emph{Proceedings of the 2022 Conference on Empirical Methods in
  Natural Language Processing}, pages 6365--6379, 2022.

\bibitem[Shi et~al.(2024)Shi, Hu, Shang, Feng, Liu, and Feng]{shi2024pose}
Tongkai Shi, Lianyu Hu, Fanhua Shang, Jichao Feng, Peidong Liu, and Wei Feng.
\newblock Pose-guided fine-grained sign language video generation.
\newblock In \emph{European Conference on Computer Vision}, pages 392--409.
  Springer, 2024.

\bibitem[Stoll et~al.(2018)Stoll, Camg{\"o}z, Hadfield, and
  Bowden]{stoll2018sign}
Stephanie Stoll, Necati~Cihan Camg{\"o}z, Simon Hadfield, and Richard Bowden.
\newblock Sign language production using neural machine translation and
  generative adversarial networks.
\newblock In \emph{Proceedings of the 29th British Machine Vision Conference
  (BMVC 2018)}. British Machine Vision Association, 2018.

\bibitem[Stoll et~al.(2022)Stoll, Mustafa, and Guillemaut]{stoll2022there}
Stephanie Stoll, Armin Mustafa, and Jean-Yves Guillemaut.
\newblock There and back again: 3d sign language generation from text using
  back-translation.
\newblock In \emph{2022 International Conference on 3D Vision (3DV)}, pages
  187--196. IEEE, 2022.

\bibitem[Tang et~al.(2022)Tang, Hong, Guo, and Wang]{tang2022gloss}
Shengeng Tang, Richang Hong, Dan Guo, and Meng Wang.
\newblock Gloss semantic-enhanced network with online back-translation for sign
  language production.
\newblock In \emph{Proceedings of the 30th ACM International Conference on
  Multimedia}, pages 5630--5638, 2022.

\bibitem[Tang et~al.(2025)Tang, Xue, Wu, Wang, and Hong]{tang2025gloss}
Shengeng Tang, Feng Xue, Jingjing Wu, Shuo Wang, and Richang Hong.
\newblock Gloss-driven conditional diffusion models for sign language
  production.
\newblock \emph{ACM Transactions on Multimedia Computing, Communications and
  Applications}, 21\penalty0 (4):\penalty0 1--17, 2025.

\bibitem[Unterthiner et~al.(2018)Unterthiner, Van~Steenkiste, Kurach, Marinier,
  Michalski, and Gelly]{unterthiner2018towards}
Thomas Unterthiner, Sjoerd Van~Steenkiste, Karol Kurach, Raphael Marinier,
  Marcin Michalski, and Sylvain Gelly.
\newblock Towards accurate generative models of video: A new metric \&
  challenges.
\newblock \emph{arXiv preprint arXiv:1812.01717}, 2018.

\bibitem[Villegas et~al.(2022)Villegas, Babaeizadeh, Kindermans, Moraldo,
  Zhang, Saffar, Castro, Kunze, and Erhan]{Villegas2022Phenaki}
Ruben Villegas, Mohammad Babaeizadeh, Pieter-Jan Kindermans, Hernan Moraldo,
  Han Zhang, Mohammad~Taghi Saffar, Santiago Castro, Julius Kunze, and Dumitru
  Erhan.
\newblock Phenaki: Variable length video generation from open domain textual
  description.
\newblock \emph{arXiv preprint arXiv:2210.02399}, 2022.

\bibitem[Wan et~al.(2025)Wan, Wang, Ai, Wen, Mao, Xie, Chen, Yu, Zhao, Yang,
  et~al.]{wan2025wan}
Team Wan, Ang Wang, Baole Ai, Bin Wen, Chaojie Mao, Chen-Wei Xie, Di~Chen,
  Feiwu Yu, Haiming Zhao, Jianxiao Yang, et~al.
\newblock Wan: Open and advanced large-scale video generative models.
\newblock \emph{arXiv preprint arXiv:2503.20314}, 2025.

\bibitem[Wang et~al.(2024)Wang, Jiang, Yuan, Peng, Wu, and
  Jiang]{Wang2024OmniTokenizer}
Junke Wang, Yi~Jiang, Zehuan Yuan, Bingyue Peng, Zuxuan Wu, and Yu-Gang Jiang.
\newblock Omnitokenizer: A joint image-video tokenizer for visual generation.
\newblock \emph{Advances in Neural Information Processing Systems},
  37:\penalty0 28281--28295, 2024.

\bibitem[Wei and Chen(2023)]{wei2023improving}
Fangyun Wei and Yutong Chen.
\newblock Improving continuous sign language recognition with cross-lingual
  signs.
\newblock In \emph{Proceedings of the IEEE/CVF International Conference on
  Computer Vision}, pages 23612--23621, 2023.

\bibitem[Xie et~al.(2024)Xie, Zhang, Taiying, Tang, Du, and Li]{xie2024g2p}
Pan Xie, Qipeng Zhang, Peng Taiying, Hao Tang, Yao Du, and Zexian Li.
\newblock G2p-ddm: Generating sign pose sequence from gloss sequence with
  discrete diffusion model.
\newblock In \emph{Proceedings of the AAAI Conference on Artificial
  Intelligence}, volume~38, pages 6234--6242, 2024.

\bibitem[Yan et~al.(2021)Yan, Zhang, Abbeel, and Srinivas]{yan2021videogpt}
Wilson Yan, Yunzhi Zhang, Pieter Abbeel, and Aravind Srinivas.
\newblock Videogpt: Video generation using vq-vae and transformers.
\newblock \emph{arXiv preprint arXiv:2104.10157}, 2021.

\bibitem[Yang et~al.(2024)Yang, Teng, Zheng, Ding, Huang, Xu, Yang, Hong,
  Zhang, Feng, et~al.]{yang2024cogvideox}
Zhuoyi Yang, Jiayan Teng, Wendi Zheng, Ming Ding, Shiyu Huang, Jiazheng Xu,
  Yuanming Yang, Wenyi Hong, Xiaohan Zhang, Guanyu Feng, et~al.
\newblock Cogvideox: Text-to-video diffusion models with an expert transformer.
\newblock \emph{arXiv preprint arXiv:2408.06072}, 2024.

\bibitem[Yin et~al.(2023)Yin, Zhong, Tang, Jin, Jin, and Zhao]{yin2023gloss}
Aoxiong Yin, Tianyun Zhong, Li~Tang, Weike Jin, Tao Jin, and Zhou Zhao.
\newblock Gloss attention for gloss-free sign language translation.
\newblock In \emph{Proceedings of the IEEE/CVF conference on computer vision
  and pattern recognition}, pages 2551--2562, 2023.

\bibitem[Zhou et~al.(2021)Zhou, Zhou, Qi, Pu, and Li]{zhou2021improving}
Hao Zhou, Wengang Zhou, Weizhen Qi, Junfu Pu, and Houqiang Li.
\newblock Improving sign language translation with monolingual data by sign
  back-translation.
\newblock In \emph{Proceedings of the IEEE/CVF conference on computer vision
  and pattern recognition}, pages 1316--1325, 2021.

\bibitem[Zuo et~al.(2025)Zuo, Potamias, Ververas, Deng, and
  Zafeiriou]{zuo2025signs}
Ronglai Zuo, Rolandos~Alexandros Potamias, Evangelos Ververas, Jiankang Deng,
  and Stefanos Zafeiriou.
\newblock Signs as tokens: A retrieval-enhanced multilingual sign language
  generator.
\newblock In \emph{Proceedings of the IEEE/CVF International Conference on
  Computer Vision}, pages 23806--23816, 2025.

\end{thebibliography}
